\documentclass[a4paper,fleqn]{cas-dc}

\usepackage[authoryear]{natbib}

\usepackage{graphicx}
\usepackage{amsmath,amssymb}
\usepackage{booktabs}
\usepackage{tabularx}
\usepackage{pdflscape}
\usepackage{array}
\newcolumntype{Y}{>{\raggedright\arraybackslash}X}
\usepackage{multirow}
\usepackage{siunitx}
\usepackage{url}
\usepackage[section]{placeins}
\usepackage{xurl}        
\usepackage{microtype}   
\usepackage{placeins}    
\usepackage{float}
\usepackage{needspace}
\newcolumntype{Y}{>{\raggedright\arraybackslash}X}
\begin{document}

\let\WriteBookmarks\relax
\def\floatpagepagefraction{1}
\def\textpagefraction{.001}


\shorttitle{CRIS: cross-plane self-supervised isotropic restoration}
\shortauthors{A. Ahituv et al.}

\title[mode=title]{CRIS: Cross-Plane Self-Supervised Isotropic Restoration for Anisotropic Volumetric Imaging Across Modalities}


\author[1]{Adi Ahituv}[orcid=0009-0009-6332-7442]
\cormark[1]
\ead{adihat3554@gmail.com}
\credit{Conceptualization, Methodology, Software, Formal analysis, Investigation, Visualization, Writing - original draft}

\author[2]{Anat Ilivitzki}[orcid=0000-0001-7770-4194]
\credit{Data curation, Validation, Investigation, Writing - review \& editing}

\author[3,4]{Moti Freiman}[orcid=0000-0003-1083-1548]
\credit{Supervision, Project administration, Funding acquisition, Writing - review \& editing}

\cortext[1]{Corresponding author}

\affiliation[1]{organization={Faculty of Data and Decision Sciences, Technion -- Israel Institute of Technology},
                city={Haifa},
                postcode={3200003},
                country={Israel}}

\affiliation[2]{organization={Rambam Health Care Campus},
                city={Haifa},
                postcode={3109601},
                country={Israel}}

\affiliation[3]{organization={Faculty of Biomedical Engineering, Technion -- Israel Institute of Technology},
                city={Haifa},
                postcode={3200003},
                country={Israel}}

\affiliation[4]{organization={The May-Blum-Dahl MRI Research Center, Technion -- Israel Institute of Technology},
                city={Haifa},
                postcode={3200003},
                country={Israel}}

\begin{abstract}
Anisotropic volumetric acquisitions are common in clinical MRI and volume electron microscopy (vEM), where sparse through-plane sampling creates thick slices or sections that degrade orthogonal reformats and downstream analysis.

We present \textbf{CRIS}, a cross-plane self-supervised framework for isotropic restoration without paired isotropic ground truth. CRIS casts 3D restoration as 2D stripe completion on orthogonal reformats of an isotropic grid: high-resolution in-plane slices are synthetically degraded and periodically masked for training, while at inference blank slices define the isotropic grid, two orthogonal reformats are restored, and predictions are fused by multi-view averaging.

We evaluate CRIS on two MRI cohorts and two microscopy benchmarks up to $8\times$ anisotropy. On brain MRI, CRIS achieves $32.921\pm0.436$\,dB PSNR and $0.963\pm0.003$ SSIM, outperforming interpolation, ECLARE, SMORE4, SIMPLE, SA-INR, and ATME, and gives the best segmentation consistency (Dice $0.940\pm0.004$, ASSD $0.245\pm0.014$\,mm, HD99 $1.275\pm0.061$\,mm). On reference-free abdominal MRI, CRIS reduces FID/KID to 48.71/0.023, outperforming interpolation, ECLARE, SMORE4, and SIMPLE. On vEM, CRIS achieves 29.100\,dB/0.830 3D PSNR/SSIM at $4\times$ and 26.874\,dB/0.722 at $8\times$ on EPFL, and 21.935$\pm$0.437\,dB/0.696$\pm$0.024 on noisy hemibrain data. In a dedicated robustness experiment, one variable-gap CRIS model evaluated across gap factors 3--7 and coronal, axial, and sagittal degradations maintained higher PSNR/SSIM than interpolation (36.36--31.14\,dB and 0.977--0.932 vs.\ 33.07--27.85\,dB and 0.951--0.853). These results support CRIS as a modality-flexible route to isotropic restoration without paired isotropic targets or configuration-specific retraining. Code is available at \url{https://github.com/adi-hatav/CRIS}.
\end{abstract}


\maketitle

\section{Introduction}

\begin{figure*}[t]
\centering
\includegraphics[width=\textwidth]{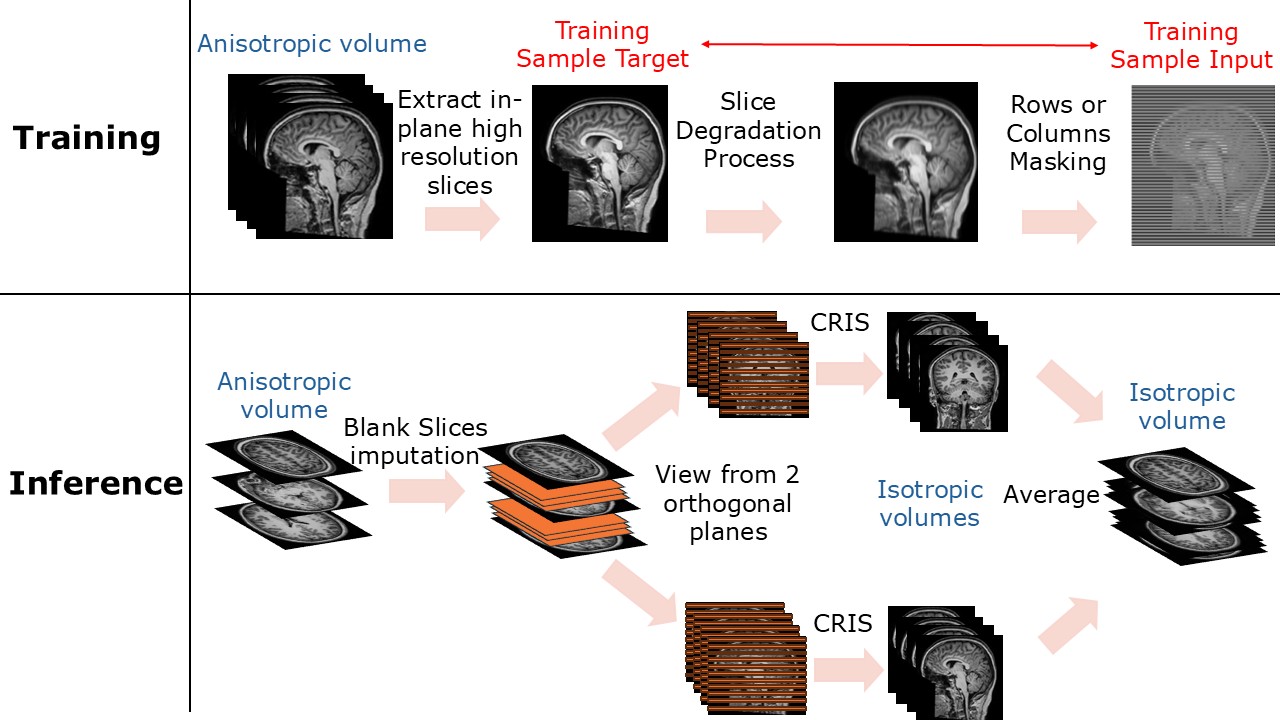}
\caption{CRIS overview during training and inference. \textbf{Training:} high-resolution in-plane slices are extracted and transformed into self-supervised input--target pairs through modality-specific degradation and periodic masking. \textbf{Inference:} blank slices are inserted to define the isotropic grid, orthogonal reformats are restored independently by the trained model, and the results are fused via multi-view averaging to obtain the final isotropic volume.}
\label{fig:cris_overview}
\end{figure*}

High-quality 3D imaging is foundational to modern medical image analysis and computational microscopy, yet the acquisition process frequently produces \emph{anisotropic} volumes: lateral (in-plane) sampling is substantially denser than sampling along the through-plane direction. In MRI, this imbalance is often an intentional compromise to reduce scan time, since acquiring thinner and denser slices generally requires longer acquisitions, while also mitigating motion artifacts and maintaining adequate signal-to-noise ratio and clinical coverage \citep{Khateri2025MRISurvey,Jack2008ADNIMethods}. However, thick slices and inter-slice gaps can introduce partial-volume effects and slice-to-slice discontinuities that degrade orthogonal reformats and can impair downstream tasks such as registration, quantitative morphometry, and segmentation \citep{Zhao2021SMORE,Remedios2023SelfSupSR,Song2024I3Net}. They also limit one of the main advantages of volumetric imaging: the ability to inspect anatomy from alternative viewing directions, where clinically relevant structural relationships may become clearer than in the acquisition plane itself. Large public repositories and multi-site studies further exacerbate the problem because real-world datasets often combine scans acquired with different through-plane spacings, slice thicknesses, and orientations \citep{DiMartino2014ABIDE,DiMartino2017ABIDEII}.

Analogous constraints appear at the nanoscale in volume electron microscopy (vEM), which is central to connectomics and ultrastructural analysis. In serial imaging pipelines, section thickness limitations and throughput/cost constraints yield markedly lower through-plane resolution than in-plane resolution, especially for large volumes \citep{PeddieCollinson2014VolumeEM,PeddieEtal2022VolumeEM}. In this setting, isotropic volumes are particularly important because many biological structures of interest, including membranes, neurites, synapses, and vesicular boundaries, extend in arbitrary 3D orientations. When the through-plane axis is sparsely sampled, these structures can appear fragmented, blurred, or topologically ambiguous in orthogonal views, complicating visual interpretation, tracing, proofreading, and downstream analysis. This is consequential in modern connectomics where datasets routinely reach terabyte scale and beyond, making re-acquisition at truly isotropic resolution expensive or infeasible at scale \citep{Scheffer2020DrosophilaConnectome,ShapsonCoe2024Petavoxel}. Computational isotropic restoration is therefore an increasingly important post-acquisition step across modalities.

A classical response to anisotropy is interpolation along the low-resolution direction (e.g., linear/cubic), including motion-aware or optical-flow variants for microscopy stacks \citep{GonzalezRuiz2022OpticalFlowFIBSEM}. While interpolation is computationally attractive, it tends to oversmooth high-frequency boundaries and can introduce implausible structures when sampling is sparse. Classical work also extended beyond direct interpolation. In MRI, early super-resolution methods combined repeated anisotropic acquisitions with sub-voxel shifts \citep{Greenspan2002MRIInterSliceSR} and explicit multi-plane acquisition models \citep{Gholipour2010MAPOrthogonalSlices,Gholipour2015FreqImageWaveletMRI} to improve through-plane resolution. Later model- and prior-based methods exploited patch recurrence, non-local self-similarity, sparse dictionaries, and auxiliary priors to recover missing high-frequency detail from low-resolution scans \citep{Manjon2010NonLocalMRIUpsampling,Rousseau2010IntermodalitySR,Rueda2013DictionaryMRI}. These approaches established the importance of acquisition-aware reconstruction and image priors, but their performance is often limited when only a single anisotropic volume is available or when priors do not transfer well across datasets and modalities. Supervised deep-learning approaches can improve fidelity, but they typically require paired isotropic ground truth (or carefully curated reference datasets) that are scarce in routine clinical imaging and in large-scale vEM \citep{Peng2020SAINT,Song2024I3Net,Iglesias2023SynthSR,Khateri2025MRISurvey}. Moreover, even when self-supervision is used, mismatches between assumed and true degradations, including scanner slice profiles, section integration, and modality-specific noise, can lead to artifacts and degraded generalization \citep{Remedios2023SelfSupSR,Khateri2025MRISurvey}. These challenges are especially important in cross-modality settings, where the degradation process is not only unknown a priori, but can differ fundamentally between imaging technologies. A useful restoration framework must therefore exploit the internal structure of anisotropic volumes, remain effective without paired isotropic targets, and retain the same central learning principle across modalities rather than being tied to MRI-specific or microscopy-specific assumptions.

We address these challenges with \textbf{CRIS}, a cross-plane self-supervised isotropic restoration framework designed to leverage a defining property of anisotropic volumes: in the original acquisition plane, the observed slices retain high in-plane resolution that can supervise restoration along the under-sampled direction. CRIS is motivated by the observation that complementary structural cues become visible when the volume is examined from multiple orientations, and that these cues can be exploited even when the target isotropic volume is never observed during training. Fig.~\ref{fig:cris_overview} illustrates how CRIS reformulates isotropic restoration as a 2D stripe-completion task on orthogonal reformats of an isotropic grid. This yields a simple learning signal: predict the missing periodic stripes from the observed content, using only the data itself. Importantly for cross-modality deployment, CRIS does \emph{not} require paired isotropic ground truth, and its central self-supervised stripe-completion formulation remains unchanged even when modality-specific degradation models or additional loss terms are introduced. The framework can therefore be adapted to modality-specific degradations without changing its core stripe-completion formulation.

Unlike prior self-supervised MRI super-resolution methods such as SMORE \citep{Zhao2021SMORE, Remedios2023SelfSupSR}, our primary novelty lies in the pipeline design and the reformulation of the restoration problem into a 2D stripe-completion task on orthogonal reformats of an isotropic grid. By explicitly inserting blank slices, we represent observed and unobserved locations directly during degradation modeling rather than hiding missing locations behind interpolation-derived proxies. This forces the network to learn conditional completion under a known geometric sampling pattern directly from the observed in-plane structure.
Additionally, CRIS enforces cross-plane consistency through multi-view restoration and fusion, targeting volumetric fidelity rather than only slice-wise appearance.

\textbf{Contributions.} This manuscript makes four contributions. First, we introduce a cross-plane self-supervised formulation for isotropic restoration that trains directly on anisotropic volumes without paired isotropic ground truth. Second, we present CRIS as a shared cross-plane learning framework with modality-specific degradation modeling and loss instantiation for both MRI and vEM, using slice-profile blur in MRI and section-integration average pooling in vEM under the same stripe-completion learning principle. Third, we demonstrate a practical robustness advantage: a single CRIS model trained with variable gap factors generalizes across multiple anisotropy severities and degradation planes, reducing the need for configuration-specific retraining. Fourth, we validate the approach on two MRI cohorts and two vEM benchmarks (EPFL FIB-SEM and FlyEM hemibrain subvolumes), reporting comprehensive quantitative metrics, qualitative orthogonal-view restorations, and downstream segmentation consistency.

\section{Related Work}

\subsection{MRI slice synthesis and volumetric super-resolution}
Deep learning for MRI super-resolution and slice synthesis has progressed rapidly, spanning supervised, weakly supervised, and self-supervised formulations \citep{Khateri2025MRISurvey}. Supervised approaches such as SAINT and I$^{3}$Net learn to map low-resolution inputs to high-resolution targets using paired supervision \citep{Peng2020SAINT,Song2024I3Net}. Other tools (e.g., SynthSR) aim to harmonize heterogeneous clinical scans into a common high-resolution representation, often enabling downstream morphometry \citep{Iglesias2023SynthSR}. While effective, supervised approaches are constrained by the availability of paired isotropic ground truth or representative high-resolution datasets, which is often unrealistic in routine clinical protocols \citep{Khateri2025MRISurvey}.

Self-supervised approaches alleviate the need for paired isotropic targets by exploiting internal redundancy. SMORE introduced a self-supervised framework for MRI super-resolution by synthesizing training pairs from the data itself and learning a slice-wise mapping from interpolated low-resolution inputs to high-resolution outputs \citep{Zhao2021SMORE}. Subsequent work highlighted the importance of accurately modeling slice gaps and acquisition physics, showing that mismatched degradations can significantly degrade reconstruction realism and downstream utility \citep{Remedios2023SelfSupSR,Lu2023CycleConsistencyTrans}. ECLARE is another closely related self-supervised MRI super-resolution method. It estimates the slice profile from the input anisotropic MR volume, trains on low- and high-resolution in-plane patch pairs from the same image, and uses field-of-view-aware resampling with anti-aliasing \citep{Remedios2026ECLARE}. We therefore include ECLARE as an additional MRI baseline in both the public brain MRI and private abdominal MRI experiments.

A key distinction lies in how unobserved through-plane information is represented and learned at inference. In SMORE, the anisotropic volume is first interpolated onto the target isotropic grid, producing a fully populated but blurred proxy volume; the network then operates slice-wise to refine this interpolated estimate \citep{Zhao2021SMORE}. In contrast, CRIS avoids introducing interpolated intensities as pseudo-observations. Instead, it explicitly models missing data by inserting blank slices at unobserved locations, yielding orthogonal reformats with structured, periodic missing stripes. A corresponding validity mask indicates which pixels are observed versus missing, and the network is trained to perform conditional completion rather than deblurring. This shift changes the learning problem from interpolation refinement to reconstruction under a known sampling pattern, where the network must infer missing through-plane content directly from observed in-plane structure. 

\subsection{Isotropic reconstruction in microscopy and vEM}
Microscopy and vEM face severe anisotropy due to physical sectioning and depth integration \citep{PeddieEtal2022VolumeEM}. Early learning-based methods explored supervised or semi-supervised isotropic reconstruction from non-isotropic EM data \citep{Heinrich2017IsotropicEMSR}. More recent work investigates self-supervised or unsupervised degradation learning to better match realistic low-resolution formation \citep{Deng2020UnsupervisedDegradation}.

Generative priors and continuous representations have become prominent. Diffusion-based approaches model a distribution over plausible reconstructions but can be computationally expensive at inference \citep{Ho2020DDPM,Song2021ScoreSDE,Pan2023DiffuseIR}. Neural implicit representations and neural fields provide resolution-agnostic decoding and can support fast inference and continuous resampling \citep{Sitzmann2020SIREN,Chen2021LIIF,Yang2026vEMINR,Troidl2026niiv}. However, implicit models can exhibit view-dependent artifacts or hallucinations under severe information loss, motivating complementary strategies that explicitly enforce cross-plane consistency.

CRIS differs in two ways that are central to this paper. First, CRIS does not require isotropic ground truth; it learns stripe completion purely from in-plane information within the same anisotropic volumes. Second, rather than decoding a continuous representation or relying on generative sampling, CRIS enforces cross-plane consistency through explicit stripe completion on orthogonal reformats followed by multi-view fusion.

\subsection{Self-supervision and internal learning without paired ground truth}
Related ideas appear in \emph{internal learning} and \emph{architectural priors}. Deep Image Prior shows that the structure of convolutional networks itself can act as a prior for restoration from a single observation \citep{Ulyanov2018DeepImagePrior}. ZSSR demonstrates zero-shot super-resolution by optimizing on the test image itself using synthetic downsampling \citep{Shocher2018ZSSR}. Blind-spot self-supervision (Noise2Void / Noise2Self) further shows how networks can learn from single noisy images under conditional independence assumptions \citep{Krull2019Noise2Void,Batson2019Noise2Self}. CRIS is related to this family of methods in that it does not require paired isotropic supervision. However, unlike architectural-prior or image-specific internal-learning methods, CRIS is a data-driven model trained across anisotropic volumes. It specifically leverages the cross-plane structure induced by anisotropy (periodic stripes in orthogonal reformats) and uses multi-view fusion to suppress view-specific artifacts.

\subsection{Cross-plane and multi-view consistency}
Exploiting multiple orientations can reduce ambiguities inherent to single-plane learning. Simultaneous multi-plane self-supervision has been shown effective for isotropic MRI restoration \citep{Benisty2026SIMPLE}. Recent single-subject multi-view MRI super-resolution with implicit neural representations further highlights the value of jointly using complementary anisotropic views together with inter-view alignment for isotropic reconstruction \citep{Kim2026SIMSMRI}. In microscopy, related multi-view strategies reconstruct orthogonal planes independently and combine predictions (e.g., via averaging) to reduce view-specific errors \citep{Yang2026vEMINR}. CRIS adopts this principle through an explicit stripe-completion formulation: it reconstructs two orthogonal reformats and fuses them by multi-view averaging, yielding a consensus isotropic volume that is typically more consistent across orientations than a single-view reconstruction.

\section{Method}

\subsection{Problem definition}
Let $V \in \mathbb{R}^{H \times W \times D}$ be an acquired anisotropic volume. Let $(r_x,r_y,r_z)$ denote the voxel spacing along the three axes in a fixed scanner/sample coordinate system, with one direction being under-sampled (the \emph{through-plane} direction). Without loss of generality, assume the through-plane spacing is $r_{\text{th}}$ and the in-plane spacing is $r_{\text{in}}$ with $r_{\text{in}} < r_{\text{th}}$. The goal is to estimate an isotropic reconstruction $\hat{V}$ sampled at $r_{\text{in}}$ along all directions.

Formally, CRIS learns parameters $\theta$ by minimizing a self-supervised reconstruction objective over high-resolution in-plane slices:
\begin{equation}
\theta^\ast =
\arg\min_{\theta}\;
\mathbb{E}_{S_{\mathrm{HR}},\,\phi}
\left[
\mathcal{L}\!\left(
f_{\theta}\!\left(\mathcal{M}_{g,\phi}\!\left(\mathcal{D}(S_{\mathrm{HR}})\right),\,M\right),
S_{\mathrm{HR}}
\right)
\right],
\end{equation}
where $\mathcal{D}$ is a modality-specific degradation operator, $\mathcal{M}_{g,\phi}$ is the periodic masking operator with gap factor $g$ and randomized phase $\phi$, $M$ is the corresponding validity mask, and $\mathcal{L}$ is the reconstruction loss defined below. At inference, the learned model is applied to orthogonal reformats of the isotropic grid and the resulting predictions are fused to obtain $\hat{V}$.

Training-pair construction in CRIS has three components:
(i) a modality-specific degradation operator $\mathcal{D}$,
(ii) periodic masking via $\mathcal{M}_{g,\phi}$ to create the striped input,
and (iii) modality-dependent intensity handling when needed to better match the target restoration distribution.
We first describe the shared formulation and then detail the MRI- and microscopy-specific instantiations.

\subsection{From 3D isotropic reconstruction to 2D stripe completion}

We begin by mapping the anisotropic volume onto an isotropic grid via explicit insertion of blank slices at unobserved locations. Let the integer gap factor be defined as
\begin{equation}
g = \operatorname{round}\!\left(\frac{r_{\mathrm{th}}}{r_{\mathrm{in}}}\right)
\end{equation}
where $r_{\text{th}}$ and $r_{\text{in}}$ denote the through-plane and in-plane resolutions, respectively. We define an insertion operator $\mathcal{I}g(\cdot)$ that places $(g-1)$ blank slices between consecutive acquired slices along the through-plane axis:
\begin{equation}
V{\text{pad}} = \mathcal{I}_g(V).
\end{equation}

From $V_{\text{pad}}$, we construct two orthogonal reformats, denoted $X^{(1)}$ and $X^{(2)}$. In these views, the inserted blanks appear as structured, periodic missing stripes within each 2D slice. This transforms the original 3D reconstruction problem into a 2D conditional completion task: given a striped slice and a corresponding binary validity mask indicating observed pixels, the model is trained to predict the missing stripe content.

For non-integer spacing ratios, we use the closest integer gap factor for blank-slice insertion and then resample the restored volume to the requested isotropic grid to account for the residual spacing difference. This practical handling is used for clinical abdominal MRI, where the through-plane-to-in-plane spacing ratio is approximately, but not exactly, 6--7.

This representation departs fundamentally from interpolation-based approaches, where missing voxels are first filled using heuristic estimates and subsequently refined. By contrast, CRIS preserves unobserved locations as explicit blanks and exposes the sampling pattern through a validity mask. As a result, the network learns to infer missing through-plane information directly from observed in-plane structure, rather than relying on interpolation-derived proxies.

\paragraph{Acquisition planes versus restoration views.}
Throughout the paper, we distinguish between the acquisition or simulated degradation plane, which defines the direction along which slices are sparse, and the restoration views, which are the two orthogonal 2D reformats processed by CRIS after blank-slice insertion. The two CRIS restoration views do not require two separate acquired scans at test time. They are generated from the same padded isotropic grid and provide complementary 2D contexts for completing the missing stripes. When multiple native acquisitions are available, as in the abdominal MRI cohort, they are used to enrich self-supervised training and validation; the reported test reconstruction is still produced from the specified anisotropic source volume.

\subsection{Self-supervised training pair construction}
CRIS trains a 2D completion network $f_\theta$ using only high-resolution in-plane slices as targets. For a sampled in-plane slice $S_{\text{HR}}$, we define the self-supervised target as
\begin{equation}
GT = S_{\text{HR}}.
\end{equation}
We then simulate through-plane information loss by applying a modality-specific degradation operator $\mathcal{D}$ followed by a periodic masking operator $\mathcal{M}_{g,\phi}$:
\begin{equation}
S_{\text{masked}} = \mathcal{M}_{g,\phi}\!\left(\mathcal{D}(GT)\right).
\end{equation}
Here, $\mathcal{M}_{g,\phi}$ retains every $g$-th row or column and masks intermediate entries; $\phi$ is a randomized phase offset to prevent periodic bias. The direction of degradation and masking (rows versus columns) is randomized per iteration so the model learns to complete stripes along both axes, matching the inference-time orthogonal reformat setting.

The network is trained to predict $\hat{S} = f_\theta(S_{\text{masked}}, M)$, where $M$ is the binary validity mask ($M=1$ for observed pixels and $M=0$ for masked pixels). Training minimizes an image-domain objective between $\hat{S}$ and $GT$, with optional modality-specific terms added when beneficial. In the MRI experiments, these include structure- and frequency-aware components such as SSIM, edge-aware loss, and focal frequency loss, whereas the microscopy setting uses a lighter objective matched to its degradation characteristics.

\paragraph{Generic degradation model.}
The operator $\mathcal{D}$ models the modality-specific mechanism by which information is lost along the through-plane direction before periodic masking. Its form depends on the imaging physics: in MRI it approximates slice-profile blurring, whereas in microscopy/vEM it approximates section-integration effects.

\paragraph{MRI-specific degradation.}
For MRI, $\mathcal{D}$ approximates slice-profile integration by applying directional 1D Gaussian blurring along a single axis, followed by periodic masking. This closely matches common synthetic anisotropy generation protocols in MRI super-resolution studies \citep{Khateri2025MRISurvey,Zhao2021SMORE,Remedios2023SelfSupSR}.

\paragraph{Microscopy-specific degradation.}
For microscopy and vEM, the physical mechanism is closer to \emph{section integration} than optical blurring. We therefore use average pooling to simulate the loss of through-plane information:
\begin{equation}
\mathcal{D}_{\text{vEM}}(GT) = \text{AvgPool}(GT;\;g),
\end{equation}
implemented as 1D average pooling along the axis corresponding to the simulated anisotropy during training or the true under-sampled axis during inference. This is aligned with recent vEM restoration protocols that model anisotropy through pooling-based degradation \citep{Yang2026vEMINR,Troidl2026niiv}.

\paragraph{MRI intensity handling during training.}
For MRI training, we apply a random intensity normalization to each sampled training slice after masking and before it is passed to the network. Specifically, we identify valid non-background pixels, randomly choose a lower clipping percentile in the range 0--12\%, clamp intensities below this percentile, and then linearly rescale the remaining range to $[-1,1]$. This acts as a simple contrast augmentation by simulating variability in low-intensity suppression and overall dynamic range. Unless stated otherwise, this normalization is applied independently to each training batch element.

\subsection{Network architecture}
CRIS uses a 2D encoder--decoder with long-range context aggregation to bridge wide missing stripes. In this work, we instantiate $f_\theta$ as a Swin-UNet-style architecture that combines local convolutions with a Swin Transformer bottleneck \citep{Liu2021SwinTransformer,Cao2021SwinUNet}. This hybrid design is well-suited to stripe completion because attention can propagate information across the stripe width while retaining local texture fidelity.
The model is trained as a two-channel input network, where the first channel contains the degraded and masked image and the second channel contains the binary validity mask. Dataset-specific patch sizes and optimization hyperparameters are summarized in Supplementary Table S2, and the released code provides the exact layer configuration used for all reported experiments.

\subsection{Objective function}
CRIS is trained with a composite reconstruction objective whose active terms can vary by modality:
\begin{equation}
\mathcal{L}(\theta) =
\alpha\,\mathcal{L}_{\mathrm{L2}}
+\beta\,\mathcal{L}_{\mathrm{SSIM}}
+\gamma\,\mathcal{L}_{\mathrm{Sobel}}
+\delta\,\mathcal{L}_{\mathrm{FFL}},
\end{equation}
where
\begin{equation}
\mathcal{L}_{\mathrm{L2}} = \|\hat{S}-GT\|_2^2
\end{equation}
enforces pointwise fidelity,
\begin{equation}
\mathcal{L}_{\mathrm{SSIM}} = 1 - \mathrm{SSIM}(\hat{S}, GT)
\end{equation}
encourages structural consistency \citep{Wang2004SSIM},
\begin{equation}
\mathcal{L}_{\mathrm{Sobel}} = \|\nabla_{\mathrm{Sobel}}\hat{S}-\nabla_{\mathrm{Sobel}}GT\|_1
\end{equation}
encourages agreement of edge magnitude maps, and $\mathcal{L}_{\mathrm{FFL}}$ denotes focal frequency loss, which emphasizes discrepancies in the frequency spectrum and is useful for suppressing periodic stripe-like artifacts \citep{Jiang2021FocalFrequencyLoss}.

The objective is instantiated differently by modality.
In MRI, all four terms are used, with the Sobel and focal-frequency components introduced later in training to sharpen restored boundaries and suppress stripe-related periodic artifacts.
In microscopy, we retain the same general objective template but disable the higher-order terms by setting
\begin{equation}
\gamma_{\text{vEM}} = 0,
\qquad
\delta_{\text{vEM}} = 0,
\end{equation}
so optimization is driven primarily by $\mathcal{L}_{\mathrm{L2}}$ and a reduced SSIM contribution.
This choice reflects the denser texture, stochastic high-frequency noise, and staining-related fluctuations in vEM, where explicitly matching gradient maps or frequency spectra can encourage reproduction of nuisance high-frequency content rather than biologically meaningful structure.

\subsection{Inference and multi-view fusion}
At inference, we apply the same insertion operator to construct an isotropic grid, reformat into two orthogonal planes, and run the 2D completion model slice-wise. This yields two restored volumes $\hat{V}^{(1)}$ and $\hat{V}^{(2)}$ after inverse reformatting back into the common 3D frame. The final isotropic volume is obtained by voxel-wise averaging:
\begin{equation}
\hat{V} = \frac{1}{2}\left(\hat{V}^{(1)} + \hat{V}^{(2)}\right).
\end{equation}
This fusion acts as a simple consensus mechanism that suppresses view-specific artifacts while preserving structures supported by both orthogonal contexts.

It is important to distinguish between the two single-view reconstructions and the fused output. The single-view reconstructions can retain mild orientation-specific advantages when inspected in the same viewing direction in which they were restored, especially for perceptual slice-based metrics. By contrast, the fused volume $\hat{V}$ is typically the most stable choice for volumetric fidelity, cross-view consistency, and downstream quantitative analysis, because it combines complementary evidence from both restored orientations and reduces directional bias.

CRIS is trained on fixed-size $H\times W$ slices, implemented here as square training patches. At inference, however, slices extracted from the isotropic grid may be larger or smaller than the training patch size along one or both dimensions. To handle this mismatch, we center-pad or crop each slice to the model input size and then restore the output back to the original slice geometry. When a slice exceeds the training patch size, we apply the model in an overlapping sliding-window manner and average the overlapping predictions to avoid seams. When a slice is smaller, we embed it into a larger canvas before inference and crop the restored result back afterward. This strategy allows CRIS to operate on arbitrarily sized reformatted slices while preserving the patch-based training regime.

The padding strategy is modality dependent. In MRI, constant padding with background values is usually acceptable because MRI training patches often contain substantial background regions outside the anatomy, so the padded intensities remain statistically plausible. In microscopy, this assumption does not hold: fields of view are densely occupied by tissue, organelles, and membranes, and artificial constant padding introduces unrealistic boundaries and distribution shifts. For this reason, microscopy inference uses reflection-style boundary handling and patterned padding that preserves the stripe geometry instead of padding with empty background.

Crucially, CRIS operates as an amortized inference framework. Unlike alternative self-supervised or internal-learning paradigms (such as SMORE4 or SA-INR) that require subject-specific optimization or fine-tuning directly on the target test volume, CRIS parameters $\theta$ are fully optimized offline. At test time, restoring an unseen anisotropic scan requires only a direct forward pass through the trained network, entirely removing the computational overhead of test-time training loops.

\subsection{Modality-specific implementation details}
\label{sec:modality_specific_details}
The CRIS formulation is shared across modalities, but several implementation choices depend on the imaging regime. We therefore summarize below the main MRI-specific and microscopy/vEM-specific details used in our experiments.

\paragraph{MRI-specific implementation details.}
For MRI, through-plane degradation is modeled with directional 1D Gaussian blurring, training uses random intensity normalization, and inference uses constant-value padding because many MRI slices contain plausible background regions outside the anatomy. MRI also uses the full four-term objective with staged activation of SSIM, focal-frequency, and Sobel losses.

\paragraph{Microscopy rotations, and training schedule.}
These choices are specific to vEM. Unlike MRI, where orientation carries anatomical meaning and arbitrary rotations may be less natural, microscopy textures are much more orientation-agnostic at the patch level. Therefore, we augment microscopy patches with additional $90^\circ$, $180^\circ$, and $270^\circ$ rotations, which increases effective data diversity without violating a canonical anatomical orientation.

Training also uses multiple inner optimization loops per epoch rather than a single pass through the loader. This repeated cycling improves stability under strong anisotropy and noisy supervision and is particularly useful for microscopy, where the degraded input-target gap is larger and the local appearance statistics are more variable.

\paragraph{vEM loss schedule and optimization.}
We use a staged training schedule to stabilize optimization. The $\mathcal{L}_{\mathrm{L2}}$ term is active from the start of training. The SSIM term is activated after an initial warm-up period, and the higher-order terms are introduced only later for MRI. In the MRI configuration, the loss weights follow
\begin{equation}
\alpha = 10,\qquad \beta = 5,\qquad \gamma = 5,\qquad \delta = 10,
\end{equation}
with staged activation of $\mathcal{L}_{\mathrm{SSIM}}$ from epoch 3, $\mathcal{L}_{\mathrm{FFL}}$ from epoch 5, and $\mathcal{L}_{\mathrm{Sobel}}$ from epoch 10. In microscopy, we retain the same schedule for the base terms, but set
\begin{equation}
\gamma = 0,\qquad \delta = 0,
\end{equation}
so Sobel and focal-frequency losses are excluded entirely. The SSIM contribution is also down-weighted relative to MRI. Optimization uses Adam with a step-based learning-rate decay factor of $0.9$, gradient clipping, and repeated sub-iterations within each epoch to improve convergence stability under noisy microscopy targets.

\paragraph{Padding and boundary handling.}
When mapping between degraded slices, masked isotropic grids, and fixed-size model inputs, boundary artifacts can arise if padding introduces values or structures absent from the training distribution. This issue is modest in MRI because many slices naturally contain extensive background. In microscopy, however, constant padding creates unrealistic borders and may shift the input distribution toward artificial background values. When boundary handling is required, microscopy inference uses reflection-aware and stripe-preserving padding rather than constant padding. In the EPFL and hemibrain experiments reported here, no explicit padding was required because all inference slice dimensions were at least $128\times128$.

\section{Datasets and Experimental Setup}

\subsection{Plane protocol}
For datasets with isotropic references, namely the public brain MRI cohort, EPFL FIB-SEM, and FlyEM hemibrain, anisotropic inputs were synthetically generated from the isotropic volumes. Therefore, coronal, axial, and sagittal in these experiments denote simulated degradation planes rather than separately acquired scans. Training and validation used anisotropic volumes generated from the corresponding train and validation splits, and all reported test metrics were computed on held-out test volumes. The primary brain MRI benchmark uses a coronal anisotropic source volume at inference, whereas the vEM benchmarks use axial through-plane degradation. The dedicated brain MRI robustness experiment is the only experiment in which the same trained CRIS model is evaluated across all three simulated degradation planes and gap factors 3--7.

\subsection{MRI datasets}
\paragraph{Public multi-site brain MRI.}
We use a public multi-site T1-weighted brain MRI cohort drawn from ABIDE and ABIDE-II \citep{DiMartino2014ABIDE,DiMartino2017ABIDEII}. The cohort contains 63 volumes originally acquired in the coronal plane at near-isotropic resolution ($1\times1\times1$\,mm or $1\times1.25\times1$\,mm). For scans with a 1.25\,mm axis, that axis was interpolated to 1\,mm to define an isotropic reference. In the primary brain MRI benchmark, anisotropic inputs were generated by simulating coronal through-plane degradation: we applied directional 1D Gaussian blurring along the acquisition axis with $\sigma=5/3$ and retained one slice every five, yielding a gap factor of $g=5$. The dataset was split into 48 training volumes, 7 validation volumes, and 8 test volumes. The exact subject identifiers used for training, validation, and testing are provided in Supplementary Table~S4.

In addition, we performed a dedicated robustness experiment on the same cohort. For this experiment, a single CRIS model was trained once using the same coronal training setup, but with randomized training-time gap factors $g\in\{3,4,5,6,7\}$ (i.e., $5\pm2$) in the self-supervised degradation process. This model was then evaluated under synthetic degradations with gap factors $3$--$7$ across all three degradation planes (coronal, axial, and sagittal). Unless stated otherwise, all other brain MRI experiments used the fixed-$g=5$ setting.

\paragraph{Private abdominal MRI.}
After Institutional Review Board (IRB) approval, we assembled a private clinical abdominal MRI cohort from Rambam Health Care Campus comprising 115 FIESTA volumes acquired on a GE 3T scanner. The cohort contains routine thick-slice clinical acquisitions in both coronal and axial orientations. Average spacing is approximately 0.76$\times$0.76$\times$4.97 mm for axial acquisitions and 0.78$\times$5.02$\times$0.78 mm for coronal acquisitions, corresponding in practice to through-plane gap factors that were mainly 6 or 7 across the cohort. We used both native coronal and axial acquisitions from the training subjects to form self-supervised training samples, and both orientations from validation subjects for model selection. For the held-out test subjects, inference and reporting were performed from the coronal anisotropic volumes only, so that the evaluation reflects restoration from a single clinical source orientation rather than fusion of two separately acquired test scans. The available native axial test acquisition was used only as an in-plane comparator for reference-free evaluation, not as an input to the reconstruction. To accommodate mild spacing heterogeneity, CRIS was trained with a variable-gap degradation setting using $g \in \{6,7\}$ rather than with a single fixed gap factor. At inference, each volume used the closest integer gap factor for blank-slice insertion, followed by resampling of the residual spacing difference to obtain the final isotropic grid. Since isotropic reference volumes are unavailable for this cohort, the evaluation is reference-free. The split is 70\% training, 15\% validation, and 15\% test.

\subsection{Microscopy and vEM datasets}
\paragraph{EPFL FIB-SEM subvolumes.}
We evaluate CRIS on the public EPFL CVLab FIB-SEM benchmark and its standard train/test subvolumes \citep{EPFLDataEM,Lucchi2013WorkingSets}. The data originates from an isotropic mouse-brain FIB-SEM stack with approximately $5\times5\times5$\,nm voxel spacing. Following the standard subvolume protocol, both the train and test volumes have dimensions $165\times768\times1024$. We simulate anisotropy by degrading the isotropic volume along the through-plane axis using average pooling, and evaluate both a moderate $4\times$ setting and a more severe $8\times$ setting.

\paragraph{FlyEM hemibrain noisy subvolumes.}
We additionally evaluate on noisy isotropic subvolumes derived from the Janelia FlyEM hemibrain project \citep{JaneliaHemibrain,Scheffer2020DrosophilaConnectome}. Native data are isotropic at approximately $8\times8\times8$\,nm resolution. Following the protocol used by NIIV, 400 randomly sampled subvolumes are constructed, each of size $128^3$, with simulated $8\times$ anisotropy induced by average pooling along the through-plane axis. The original protocol uses 350 subvolumes for training and 50 for testing. In our experiments, we held out cases 301--350 from the \texttt{hemibrain-volume-noisy-large} training portion for validation, yielding 300 training, 50 validation, and 50 test subvolumes.

\subsection{Baselines and evaluation metrics}
For both MRI and microscopy evaluations, FID and KID were computed using a pre-trained Inception-V3 backbone \citep{Szegedy2016Inception} with the final classification head replaced by an identity mapping to extract feature representations. Because FID and KID rely on feature representations learned from natural images, we use them here only as complementary perceptual-distribution metrics rather than as direct measures of anatomical or biological validity. For the abdominal MRI cohort, where isotropic references are unavailable, FID/KID should therefore be interpreted as reference-free distributional evidence rather than proof of anatomical correctness. Prior to quantitative evaluation, MRI volumes were clipped at the 99.9th percentile to remove extreme intensity outliers and subsequently normalized to the $[0, 1]$ range via min-max scaling to prevent massive contrast fluctuations. This evaluation-time normalization is separate from the random training-time intensity normalization used as augmentation during CRIS training. Microscopy data were similarly normalized using min-max scaling before all metric calculations.

\paragraph{Baseline protocol.}
All learning-based baselines were trained and selected without using the held-out isotropic test references for optimization or hyperparameter selection. When official implementations were available, we used the authors' supplied code and adapted only the data interface, spacing or gap configuration, and evaluation input/output needed to match our train, validation, and test splits. Methods that are normally optimized per subject were run per test case, whereas dataset-level methods were trained on the corresponding training split and selected on validation data. The same source orientation used for CRIS test inference was used for the corresponding baseline comparison.

\paragraph{MRI baselines.}
For brain MRI, we compare against interpolation, ECLARE \citep{Remedios2026ECLARE}, SMORE4 \citep{Zhao2021SMORE}, SIMPLE \citep{Benisty2026SIMPLE}, SA-INR \citep{Wang2024SAINR}, and ATME \citep{SolanoCarrillo2023ATME}. For abdominal MRI, we compare against interpolation, ECLARE, SMORE4, and SIMPLE. SMORE4 was run using its supplied implementation and optimized separately for each held-out coronal test volume. SIMPLE was run with the supplied code and data configuration for abdominal MRI, and with a minor configuration adaptation for brain MRI, which is outside the original target domain of SIMPLE. SA-INR was included as an adapted implicit-representation baseline: because paired isotropic supervision is unavailable in our self-supervised setting, the training and validation data supplied to SA-INR were anisotropic coronal brain MRI volumes rather than paired isotropic ground-truth volumes, and a separate SA-INR network was trained for each test case. ATME was trained separately for each acquisition or degradation plane using paired 2D samples derived from the anisotropic training volumes. Its input was generated from the anisotropic volume by linear resampling or interpolation to the target grid together with the plane-dependent stride pattern, and the target was the corresponding real acquired slice from the original anisotropic volume. The model was optimized as a conditional image-to-image translation baseline using an adversarial loss with a PatchGAN discriminator and an L1 reconstruction loss. Brain MRI metrics are PSNR, SSIM, FID, KID, GMSD, Edges 3D, and downstream brain segmentation Dice using SynthSeg robust inference \citep{Wang2004SSIM,Xue2014GMSD,Billot2023SynthSeg,Billot2023SynthSegRobust}. On the abdominal cohort, because isotropic reference volumes are unavailable, we report FID and KID by comparing restored coronal and axial slices to the corresponding original in-plane slices.

\paragraph{Microscopy/vEM baselines.}
For EPFL, we compare against bicubic interpolation, NIIV \citep{Troidl2026niiv}, and vEMINR \citep{Yang2026vEMINR}. For the EPFL $4\times$ setting only, because the available training data were limited, the degraded anisotropic axial input volume was also included as an additional self-supervised training volume, without using the isotropic test reference. For the hemibrain noisy benchmark, we compare against NIIV, nearest-neighbor interpolation, and bilinear interpolation, following the evaluation style used in the NIIV work. NIIV and vEMINR were run using the authors' supplied implementations. For EPFL, the standard degraded training and testing volumes were converted into the data format expected by NIIV, and the patch-wise NIIV outputs were reassembled into full volumes before evaluation. vEMINR was trained using the supplied implementation on the degraded EPFL volume, and we report both directional reconstructions (XZ and YZ) as well as their final merged volume. This ensures that each baseline is evaluated under its intended training and inference pipeline while using the same held-out test volumes as CRIS. For EPFL, we report 3D PSNR, 3D SSIM, GMSD, FID, and KID in the main tables; for the noisy hemibrain benchmark we additionally report CF-PSNR following NIIV \citep{Troidl2026niiv}. Lower FID/KID and GMSD indicate better restoration quality. Detailed baseline implementation notes, including the adaptations used for SMORE4, SIMPLE, SA-INR, ATME, NIIV, and vEMINR, are provided in the Supplementary Material under ``Baseline implementation details''.

\subsection{Statistical analysis}
For metrics available at the case level, statistical significance was assessed using paired two-sided Wilcoxon signed-rank tests comparing CRIS with each baseline on matched test cases. The paired difference was defined so that positive values indicate better CRIS performance, according to the metric direction: higher is better for PSNR, CF-PSNR, SSIM, and Edges 3D, whereas lower is better for GMSD, ASSD, HD99, and related error or distance metrics. Holm--Bonferroni correction was applied separately within each metric across the CRIS-versus-baseline comparisons. Distribution-level metrics such as FID and KID were reported descriptively and were not included in paired significance testing. For downstream segmentation, statistical significance was evaluated at both the volume-averaged level and the per-structure level using paired two-sided Wilcoxon signed-rank tests. Volume-averaged metrics were corrected using the Holm--Bonferroni method, while per-structure comparisons across the 32 anatomical labels were corrected using the Benjamini-Hochberg False Discovery Rate (FDR) procedure to account for multiple anatomical comparisons.

\subsection{Implementation notes and sigma calibration}
For MRI, the slice-select blur parameter $\sigma$ was selected using validation-set experiments, with the selection criterion matched to the available reference information in each cohort. In the public brain MRI benchmark, where isotropic validation references are available, validation-set GT-referenced FID and PSNR identify an optimum near $\sigma \approx 1.67$, whereas in-plane-referenced FID misleadingly favors $\sigma=0$. We therefore use the physics-informed setting $\sigma = 0.75\sqrt{g}$ for the public brain MRI experiments, which gives $\sigma \approx 1.67$ for $g=5$. In the private abdominal MRI cohort, isotropic references are unavailable and validation is necessarily reference-free; the reported abdominal results use $\sigma=0$, because this setting maximized the available in-plane validation metrics. This difference reflects the dataset-specific validation criterion rather than a change in the CRIS formulation. Supplementary Table S2 summarizes the dataset-specific CRIS configurations and selected hyperparameters across the four datasets.

\section{Results}

\subsection{Brain MRI results}
Table~\ref{tab:brain_mri_results} reports reference-based quantitative evaluation on the public brain MRI cohort. CRIS achieves the best performance across all reported fidelity and perceptual metrics, with 32.921$\pm$0.436\,dB PSNR, 0.963$\pm$0.003 SSIM, 35.039 FID, 0.013 KID, and 0.015$\pm$0.001 GMSD. It improves over interpolation, ECLARE, SMORE4, SIMPLE, SA-INR, and ATME. For all case-level metrics reported with SD, CRIS was better than every baseline on all eight matched test volumes. Using two-sided paired Wilcoxon signed-rank tests with Holm--Bonferroni correction across baseline comparisons within each metric, all CRIS-versus-baseline differences in case-level image-quality metrics remained statistically significant ($p_{\mathrm{adj}}=0.039$). Given the small paired test set ($n=8$), these significance results should be interpreted cautiously and primarily as support for the consistent paired effect sizes; nevertheless, CRIS outperformed each baseline on every matched test case for the reported case-level image-quality metrics. In downstream analysis, CRIS also achieved the strongest segmentation consistency, with the best mean Dice, ASSD, and HD99 across 32 anatomical structures. These volume-averaged segmentation improvements over both SMORE4 and interpolation were statistically significant across all three metrics ($p_{\mathrm{adj}} = 0.008$, using paired Wilcoxon signed-rank tests with Holm--Bonferroni correction). Furthermore, at the individual structure level (corrected for multiple comparisons using the Benjamini-Hochberg False Discovery Rate), CRIS yielded significantly better Dice scores than SMORE4 in 29 out of 32 evaluated anatomical regions.

\begin{table*}[pos=t]
\caption{Brain MRI quantitative image-quality results on the public multi-site cohort with simulated through-plane gap factor $g=5$. PSNR, SSIM, GMSD, and Edges 3D are reported as mean $\pm$ SD over the eight test volumes. FID and KID are distribution-level metrics and are reported without SD. Higher is better for PSNR, SSIM, and Edges 3D; lower is better for FID, KID, and GMSD.}
\label{tab:brain_mri_results}
\centering
\scriptsize
\setlength{\tabcolsep}{3pt}
\begin{tabular*}{\tblwidth}{@{} LCCCCCC @{}}
\toprule
Method & PSNR $\uparrow$ & SSIM $\uparrow$ & FID $\downarrow$ & KID $\downarrow$ & GMSD $\downarrow$ & Edges 3D $\uparrow$ \\
\midrule
CRIS & \textbf{32.921 $\pm$ 0.436} & \textbf{0.963 $\pm$ 0.003} & \textbf{35.04} & \textbf{0.013} & \textbf{0.0152 $\pm$ 0.001} & \textbf{0.949 $\pm$ 0.004} \\
SMORE4 & 31.963 $\pm$ 0.455 & 0.953 $\pm$ 0.004 & 45.59 & 0.020 & 0.0210 $\pm$ 0.001 & 0.934 $\pm$ 0.005 \\
ECLARE & 31.364 $\pm$ 0.522 & 0.946 $\pm$ 0.005 & 51.46 & 0.026 & 0.0235 $\pm$ 0.002 & 0.926 $\pm$ 0.007 \\
ATME & 25.685 $\pm$ 0.702 & 0.877 $\pm$ 0.008 & 91.58 & 0.059 & 0.0620 $\pm$ 0.0040 & 0.843 $\pm$ 0.009 \\
SIMPLE & 20.192 $\pm$ 0.423 & 0.706 $\pm$ 0.010 & 126.27 & 0.084 & 0.0890 $\pm$ 0.0030 & 0.691 $\pm$ 0.013 \\
SA-INR & 20.703 $\pm$ 0.963 & 0.750 $\pm$ 0.019 & 101.78 & 0.066 & 0.1000 $\pm$ 0.0060 & 0.704 $\pm$ 0.019 \\
Interpolation & 28.784 $\pm$ 0.467 & 0.908 $\pm$ 0.007 & 102.07 & 0.067 & 0.0440 $\pm$ 0.0020 & 0.867 $\pm$ 0.009 \\
\bottomrule
\end{tabular*}
\end{table*}

To better understand the role of fusion, we separately evaluated the two single-view reconstructions and the fused output. The fused output achieved the best mean PSNR and SSIM across the three evaluation directions, supporting its use as the default volume for quantitative 3D analysis. The single-view reconstructions, however, remained slightly more favorable in some view-matched perceptual comparisons, indicating mild orientation-specific specialization. This behavior is consistent with the design of CRIS: single-view restoration can be visually sharp in a matched viewing direction, whereas fusion provides the strongest overall volumetric consistency.
Table~\ref{tab:brain_fusion_analysis} summarizes the comparison between the single-view reconstructions and the fused output. Overall, fusion provided the strongest voxel-space fidelity, whereas single-view reconstructions retained minor advantages in a few plane-matched perceptual comparisons.

\begin{table*}[t]
\caption{Plane-specific analysis of the brain MRI CRIS outputs. ``Fused'' denotes the voxel-wise average of the two single-view reconstructions. PSNR and SSIM are reported as mean $\pm$ SD across test cases. Higher is better for PSNR and SSIM; lower is better for FID and KID.}
\label{tab:brain_fusion_analysis}
\centering
\scriptsize
\setlength{\tabcolsep}{2pt}
\resizebox{\textwidth}{!}{%
\begin{tabular}{lcccccccccccccccc}
\toprule
& \multicolumn{4}{c}{Coronal plane} & \multicolumn{4}{c}{Axial plane} & \multicolumn{4}{c}{Sagittal plane} & \multicolumn{4}{c}{Mean over planes} \\
\cmidrule(lr){2-5}\cmidrule(lr){6-9}\cmidrule(lr){10-13}\cmidrule(lr){14-17}
Output & PSNR $\uparrow$ & SSIM $\uparrow$ & FID $\downarrow$ & KID $\downarrow$
& PSNR $\uparrow$ & SSIM $\uparrow$ & FID $\downarrow$ & KID $\downarrow$
& PSNR $\uparrow$ & SSIM $\uparrow$ & FID $\downarrow$ & KID $\downarrow$
& PSNR $\uparrow$ & SSIM $\uparrow$ & FID $\downarrow$ & KID $\downarrow$ \\
\midrule
Fused
& \textbf{33.032 $\pm$ 0.555} & \textbf{0.967 $\pm$ 0.003} & \textbf{17.71} & 0.001
& \textbf{32.262 $\pm$ 0.601} & \textbf{0.965 $\pm$ 0.002} & 35.87 & 0.012
& \textbf{33.470 $\pm$ 0.567} & \textbf{0.957 $\pm$ 0.003} & 51.53 & 0.026
& \textbf{32.921 $\pm$ 0.436} & \textbf{0.963 $\pm$ 0.003} & 35.04 & 0.013 \\

Axial-view
& 32.526 $\pm$ 0.594 & 0.964 $\pm$ 0.004 & 17.800 & \textbf{0.001}
& 31.777 $\pm$ 0.590 & 0.962 $\pm$ 0.002 & \textbf{33.941} & \textbf{0.010}
& 33.047 $\pm$ 0.555 & 0.953 $\pm$ 0.004 & 60.274 & 0.035
& 32.450 $\pm$ 0.437 & 0.959 $\pm$ 0.003 & 37.338 & 0.016 \\

Sagittal-view
& 32.587 $\pm$ 0.522 & 0.964 $\pm$ 0.003 & 18.140 & 0.002
& 31.793 $\pm$ 0.588 & 0.962 $\pm$ 0.002 & 38.718 & 0.014
& 33.042 $\pm$ 0.581 & 0.953 $\pm$ 0.003 & \textbf{47.786} & \textbf{0.020}
& 32.474 $\pm$ 0.432 & 0.960 $\pm$ 0.003 & \textbf{34.882} & \textbf{0.012} \\
\bottomrule
\end{tabular}
}
\end{table*}

\begin{figure*}[t]
\centering
\includegraphics[width=\textwidth]{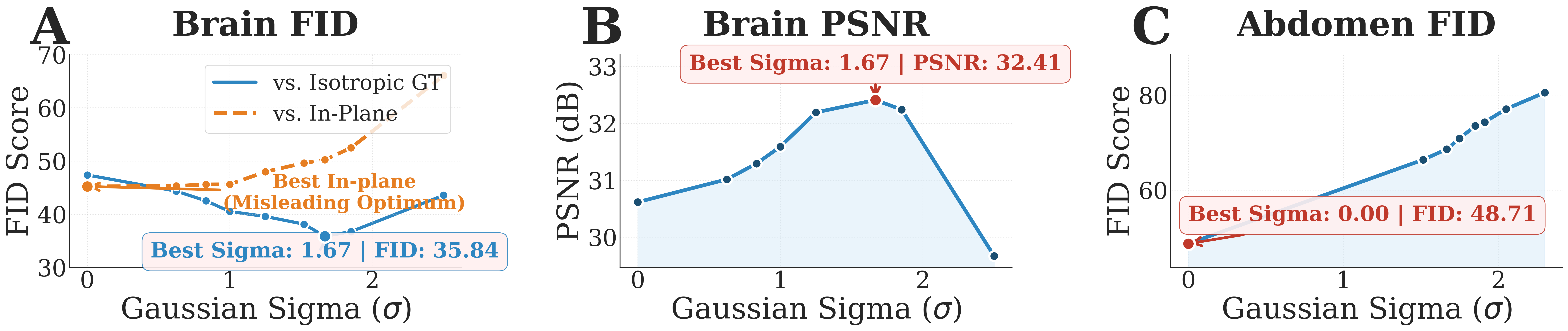}
\caption{Validation-set $\sigma$ calibration for MRI slice-select simulation. (A) On the brain MRI validation set, GT-referenced FID attains its best value near $\sigma\approx1.67$, whereas in-plane-referenced FID misleadingly favors $\sigma=0$. (B) Brain test-set PSNR also peaks near $\sigma\approx1.67$. (C) On the abdominal MRI validation set, in-plane-referenced FID again favors $\sigma=0$, showing the same bias. These validation trends motivate the physics-informed choice of $\sigma$ and illustrate the sigma paradox: purely in-plane evaluation can prefer unrealistic degradations that do not yield the most anatomically plausible restored volumes.}
\label{fig:sigma_paradox}
\end{figure*}

Figure~\ref{fig:sigma_paradox} summarizes the validation-set MRI blur-calibration analysis used for selecting $\sigma$.

\begin{figure*}[t]
\centering
\includegraphics[width=\textwidth]{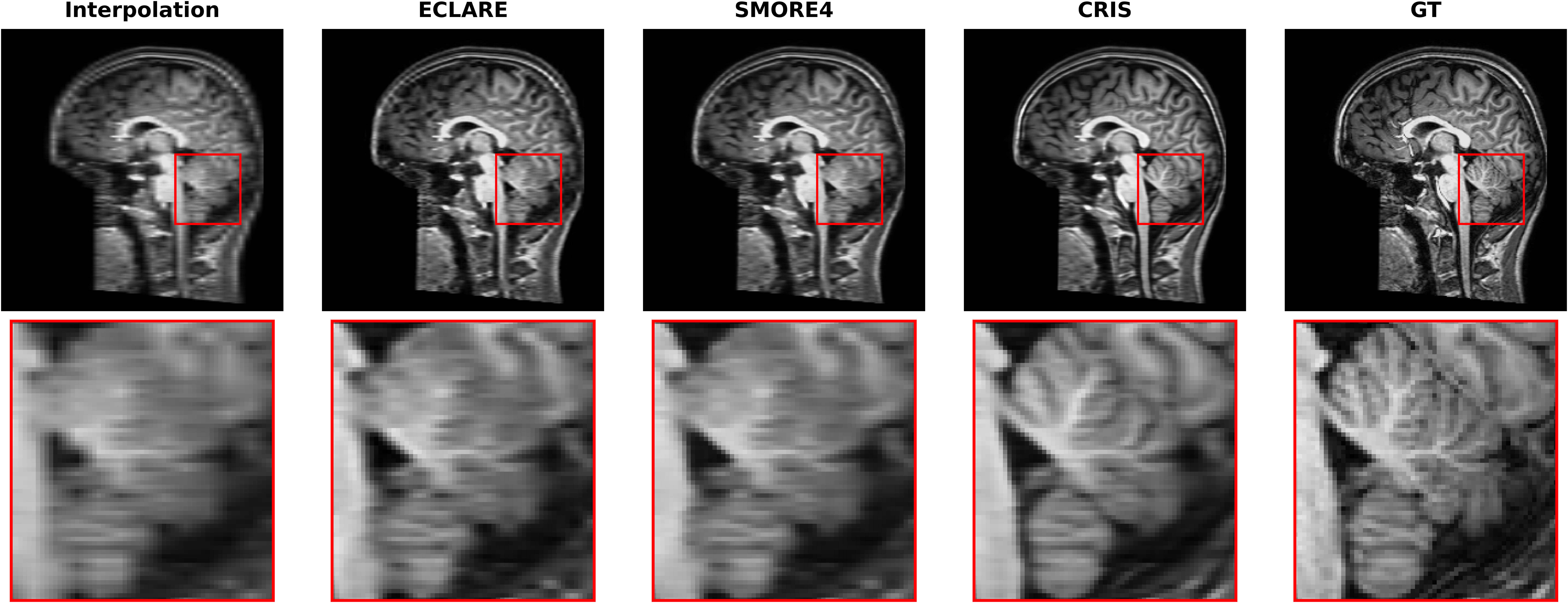}
\vspace{2mm}
\includegraphics[width=\textwidth]{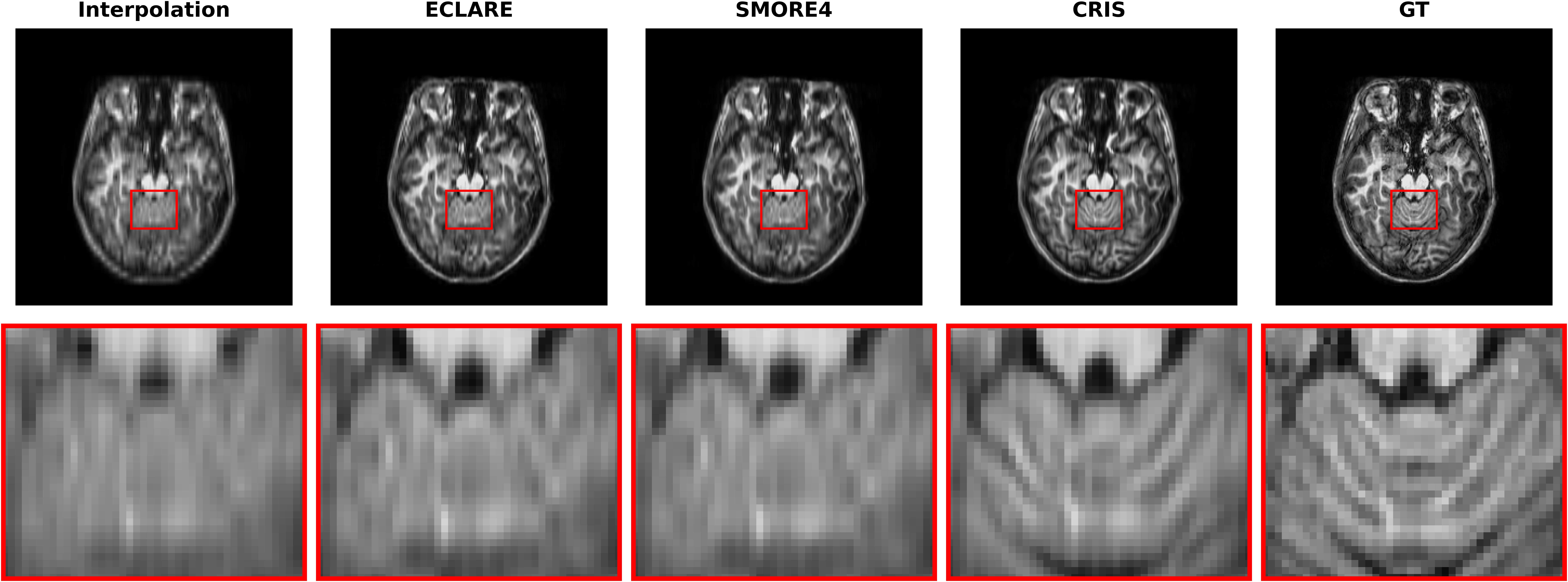}
\caption{Brain MRI qualitative comparison showing zoomed-in regions of interest across sagittal and axial planes from an anisotropic coronal source volume with gap factor $g=5$. Columns are ordered left-to-right as Interpolation, ECLARE, SMORE4, CRIS, and isotropic ground truth (GT).}
\label{fig:brain_visual_results}
\end{figure*}

Figure~\ref{fig:brain_visual_results} shows qualitative brain MRI results across sagittal and axial views. CRIS reduces staircase artifacts and yields more anatomically coherent orthogonal reformats than interpolation, ECLARE, and SMORE4, while better approaching the isotropic reference. This qualitative improvement is consistent with the gains in PSNR, perceptual metrics, and segmentation Dice reported in Table~\ref{tab:brain_mri_results}.

Beyond image fidelity, we evaluated downstream segmentation consistency across 32 anatomical structures. We first segmented the isotropic reference volume and then applied the same segmentation pipeline to each reconstructed anisotropic-to-isotropic volume, followed by structure-wise comparison using Dice, ASSD, and HD99. As summarized in Table~\ref{tab:brain_segmentation_summary}, CRIS achieved the best mean segmentation performance, reaching $0.940\pm0.004$ Dice, $0.245\pm0.014$\,mm ASSD, and $1.275\pm0.061$\,mm HD99, compared with $0.923\pm0.005$ / $0.311\pm0.019$\,mm / $1.550\pm0.090$\,mm for SMORE4 and $0.911\pm0.004$ / $0.364\pm0.022$\,mm / $1.825\pm0.098$\,mm for interpolation. The gains were especially pronounced in representative structures such as cerebral white matter, inferior lateral ventricles, and the 4th ventricle, indicating that improved restoration fidelity translates into more anatomically consistent downstream analysis. Supplementary Table~S1 summarizes the per-structure Dice, ASSD, and HD99 results across all 32 anatomical labels.

\begin{table}[t]
\caption{Brain MRI downstream segmentation summary over 32 anatomical structures. Values are reported as mean $\pm$ SD over the eight test volumes after averaging each metric across the 32 structures within each volume. The isotropic reference volume was segmented first, and the resulting segmentations were compared against segmentations obtained from the reconstructed volumes. Higher is better for Dice; lower is better for ASSD and HD99. Volume-averaged improvements of CRIS over both baselines are statistically significant across all three metrics ($p_{\mathrm{adj}} < 0.01$).}
\label{tab:brain_segmentation_summary}
\centering
\begin{tabular*}{\columnwidth}{@{\extracolsep{\fill}}lccc@{}}
\toprule
Method & Dice $\uparrow$ & ASSD $\downarrow$ & HD99 $\downarrow$ \\
\midrule
CRIS & \textbf{0.940 $\pm$ 0.004} & \textbf{0.245 $\pm$ 0.014} & \textbf{1.275 $\pm$ 0.061} \\
SMORE4 & 0.923 $\pm$ 0.005 & 0.311 $\pm$ 0.019 & 1.550 $\pm$ 0.090 \\
Interpolation & 0.911 $\pm$ 0.004 & 0.364 $\pm$ 0.022 & 1.825 $\pm$ 0.098 \\
\bottomrule
\end{tabular*}
\end{table}

For example, CRIS improved Dice over interpolation in left/right cerebral white matter (0.953/0.952 vs.\ 0.926/0.926), left/right inferior lateral ventricles (0.837/0.845 vs.\ 0.771/0.762), and the 4th ventricle (0.928 vs.\ 0.872).

\paragraph{Computational efficiency and runtime analysis.}
Offline training of the baseline brain-MRI CRIS model was performed on a single NVIDIA RTX A6000 GPU. A full training epoch with the complete objective required approximately 477 s, with only a minor reduction of several seconds when optimizing with the $\mathcal{L}_{2}$ term alone. To evaluate inference efficiency, the trained model was deployed on the same GPU and applied to the public brain MRI test set. Each anisotropic input volume, with native dimensions of $256 \times 45 \times 256$ voxels, was expanded along the coronal acquisition axis ($g=5$) to $256 \times 221 \times 256$ voxels after blank-slice insertion. The resulting axial and sagittal reformatted slices were center-padded to $256 \times 256$ pixels and processed slice-wise. Structural validity masks were used to skip empty imputed slices, thereby avoiding redundant network evaluations. GPU-only inference averaged $8.1 \pm 0.4$ s per volume. Including disk I/O, structural padding, orthogonal reformatting, and voxel-wise consensus fusion, end-to-end restoration required $14.8 \pm 0.5$ s per volume and processed the full eight-subject test set in under 2 min.

\subsection{Brain MRI robustness across gap factors and degradation planes}

\begin{figure*}[t]
\centering
\includegraphics[width=\textwidth]{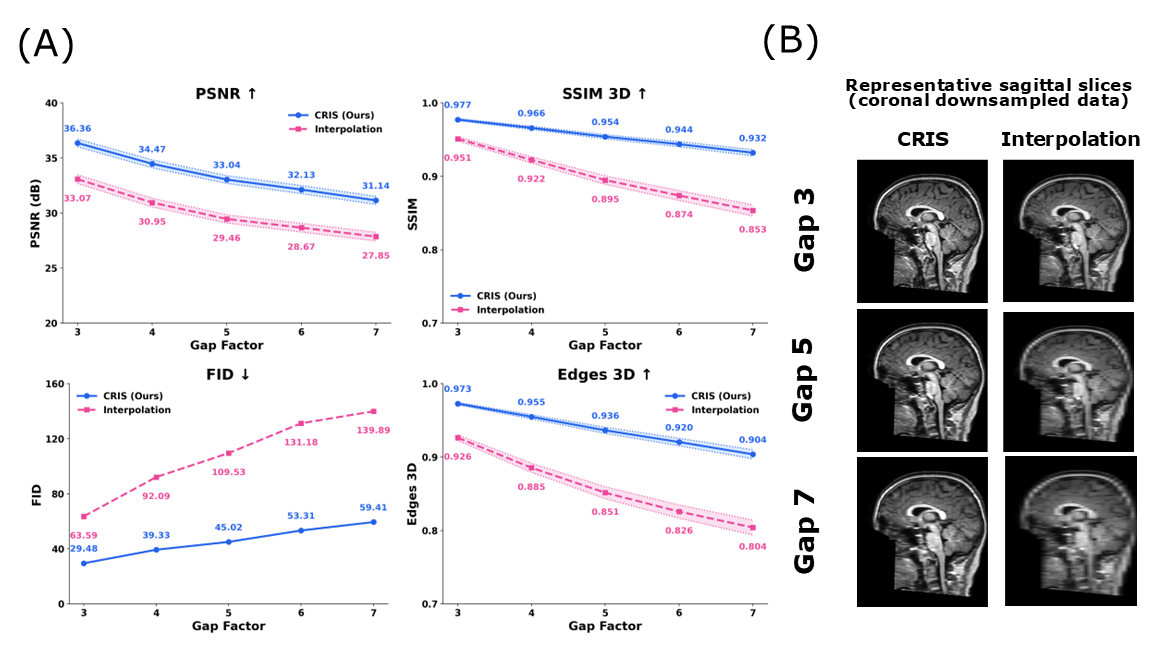}
\caption{Brain MRI robustness across anisotropy severity and degradation orientation. A single CRIS model was trained once using coronal synthetic degradation with randomized training-time gap factors $g\in\{3,4,5,6,7\}$ in the self-supervised degradation process. The trained model was then evaluated under synthetic degradations with gap factors $3$--$7$ across coronal, axial, and sagittal degradation planes. (A) Quantitative metrics averaged across the three degradation planes. (B) Representative sagittal-view slices under coronal degradation for selected gap factors. CRIS consistently outperforms interpolation and degrades gracefully as the gap increases.}
\label{fig:brain_gap_robustness}
\end{figure*}

To assess robustness beyond the primary coronal $g=5$ benchmark, we trained a single additional CRIS model on the public brain MRI cohort using coronal synthetic degradation with randomized training-time gap factors $g\in\{3,4,5,6,7\}$ in the self-supervised degradation process. We then evaluated this same trained model, without retraining or configuration-specific tuning, on test volumes synthetically degraded with gap factors $g\in\{3,4,5,6,7\}$ in each of the coronal, axial, and sagittal planes. Figure~\ref{fig:brain_gap_robustness} summarizes the results.

This experiment is intentionally compared only against interpolation. Unlike interpolation, each additional learning-based baseline would require separate retraining, degradation calibration, and hyperparameter tuning for each gap factor and degradation plane to ensure a fair comparison. The purpose of this experiment is therefore not to repeat the full benchmark comparison from Table~\ref{tab:brain_mri_results}, but to test a distinct practical question: whether a single CRIS model can generalize across anisotropy severities and acquisition orientations without per-configuration retraining.

Averaged across degradation planes, CRIS consistently outperformed interpolation at every gap factor. PSNR decreased gradually from 36.36\,dB at $g=3$ to 31.14\,dB at $g=7$, whereas interpolation decreased from 33.07\,dB to 27.85\,dB. Similarly, 3D SSIM decreased from 0.977 to 0.932 for CRIS versus 0.951 to 0.853 for interpolation, and Edges 3D decreased from 0.973 to 0.904 for CRIS versus 0.926 to 0.804 for interpolation. Although performance declined as the gap increased, CRIS degraded gracefully and maintained a clear margin over interpolation, supporting the claim that its learned cross-plane completion prior is not restricted to a single acquisition geometry. FID likewise remained substantially lower across the full range, increasing from 29.48 to 59.41 for CRIS compared with 63.59 to 139.89 for interpolation.

Representative qualitative results show the same trend: CRIS better preserves anatomical continuity and suppresses cross-plane blurring and staircasing artifacts across all three degradation planes, even at the more challenging gap factors. Full per-configuration quantitative results, including GMSD and KID, are reported in Supplementary Table~S3, and expanded qualitative cross-plane comparisons are provided in Supplementary Fig.~S1. Importantly, these results were obtained with one trained CRIS model rather than separate models for each gap factor or degradation plane.

\subsection{Brain MRI component ablation study}

Separate from the robustness experiment above, we performed a component ablation study on the main brain MRI setting to assess the contribution of key design choices. Starting from the baseline CRIS configuration, we varied components to assess their individual contributions. This included ablating individual and combined loss terms (training with all losses active from the start, removing the Sobel loss, removing both Sobel and focal-frequency losses, and training with only the L2 loss). We also evaluated the removal of the validity-mask input channel and random intensity normalization, altering the dimensionality of the degradation kernel, and the application of spatial rotation augmentations ($0^{\circ}, 90^{\circ}, 180^{\circ}$, and $270^{\circ}$) during training. Gap randomization is analyzed separately in the dedicated robustness experiment because it changes the training distribution and is intended to test single-model generalization across anisotropy severities and degradation planes. Table~\ref{tab:brain_ablation} summarizes the results.

The baseline configuration (with staged loss scheduling) achieved the best case-level fidelity and structural consistency on the held-out brain MRI test set, yielding the highest PSNR, SSIM, and Edges 3D, alongside the lowest GMSD. Activating all losses from the start reduced performance relative to the staged schedule, supporting the use of a warm-up period before the higher-order losses are introduced. Removing the Sobel loss, removing both Sobel and focal-frequency losses, or relying only on the L2 loss reduced structural fidelity and edge preservation. Removing the focal-frequency loss individually produced a small distribution-level improvement in FID and KID, but reduced voxel-based fidelity and structural metrics, indicating a trade-off between distributional alignment and anatomical consistency. Adding rotation augmentation also reduced performance, consistent with the strongly canonical orientation of brain anatomy. Disabling random training-time intensity normalization decreased performance across all metrics, and replacing the 1D degradation kernel with a 2D Gaussian kernel caused the largest degradation, supporting the use of direction-specific slice-profile simulation in MRI.

\begin{table*}[t]
\caption{Brain MRI component ablation study on the held-out brain MRI test set of the public multi-site cohort. Starting from the baseline CRIS configuration, we modify selected design choices or training schedules. PSNR, SSIM, GMSD, and Edges 3D are reported as mean $\pm$ SD over the eight test volumes. FID and KID are distribution-level metrics and are reported without SD. FID is rounded to two decimal places; GMSD is rounded to four decimal places; PSNR, SSIM, KID, and Edges 3D are rounded to three decimal places. Higher is better for PSNR, SSIM, and Edges 3D; lower is better for FID, KID, and GMSD. Best values in each column are shown in bold.}
\label{tab:brain_ablation}
\centering
\scriptsize
\setlength{\tabcolsep}{3pt}
\resizebox{\textwidth}{!}{%
\begin{tabular}{lcccccc}
\toprule
Variant & PSNR $\uparrow$ & SSIM $\uparrow$ & FID $\downarrow$ & KID $\downarrow$ & GMSD $\downarrow$ & Edges 3D $\uparrow$ \\
\midrule
Baseline model & \textbf{32.921 $\pm$ 0.436} & \textbf{0.963 $\pm$ 0.003} & 35.04 & 0.013 & \textbf{0.0152 $\pm$ 0.0007} & \textbf{0.950 $\pm$ 0.004} \\
No focal-frequency loss & 32.767 $\pm$ 0.440 & 0.963 $\pm$ 0.003 & \textbf{34.77} & \textbf{0.013} & 0.0157 $\pm$ 0.0008 & 0.949 $\pm$ 0.004 \\
All losses from start & 32.746 $\pm$ 0.431 & 0.962 $\pm$ 0.003 & 36.05 & 0.013 & 0.0156 $\pm$ 0.0007 & 0.947 $\pm$ 0.004 \\
No Sobel loss & 32.646 $\pm$ 0.434 & 0.962 $\pm$ 0.003 & 37.26 & 0.014 & 0.0156 $\pm$ 0.0007 & 0.947 $\pm$ 0.004 \\
No Sobel or focal-frequency loss & 32.746 $\pm$ 0.445 & 0.962 $\pm$ 0.003 & 37.02 & 0.014 & 0.0157 $\pm$ 0.0007 & 0.947 $\pm$ 0.004 \\
Only L2 loss & 32.681 $\pm$ 0.418 & 0.958 $\pm$ 0.003 & 39.08 & 0.016 & 0.0160 $\pm$ 0.0007 & 0.943 $\pm$ 0.005 \\
Add rotation augmentation & 32.790 $\pm$ 0.445 & 0.963 $\pm$ 0.003 & 36.51 & 0.014 & 0.0159 $\pm$ 0.0008 & 0.949 $\pm$ 0.004 \\
No random intensity normalization & 32.616 $\pm$ 0.442 & 0.962 $\pm$ 0.003 & 38.23 & 0.016 & 0.0167 $\pm$ 0.0008 & 0.947 $\pm$ 0.004 \\
No extra mask channel & 32.715 $\pm$ 0.429 & 0.963 $\pm$ 0.003 & 36.10 & 0.014 & 0.0158 $\pm$ 0.0008 & 0.949 $\pm$ 0.004 \\
2D kernel & 27.300 $\pm$ 0.427 & 0.918 $\pm$ 0.006 & 66.07 & 0.035 & 0.0324 $\pm$ 0.0011 & 0.895 $\pm$ 0.008 \\
\bottomrule
\end{tabular}
}
\end{table*}

\subsection{Abdominal MRI results}
Table~\ref{tab:abdominal_mri_results} reports reference-free evaluation on the private abdominal MRI cohort. For the held-out test subjects, the input to CRIS and all baselines was the coronal anisotropic volume. The native axial acquisition was not used as an input for reconstruction; it was used only as an additional in-plane comparator for reference-free evaluation. Since no isotropic reference is available, evaluation compares restored coronal and axial slices against the corresponding original in-plane slices using perceptual feature-space metrics. CRIS achieves the best performance, reducing FID to 48.71 and KID to 0.023, substantially outperforming interpolation, ECLARE, SIMPLE, and SMORE4. These gains indicate improved cross-plane anatomical consistency and fewer view-specific artifacts in the restored volume.

\begin{table}[pos=t]
\caption{Reference-free abdominal MRI results on the private Rambam FIESTA cohort with typical through-plane gap factor $g\approx6$. Evaluation compares restored orthogonal slices against the corresponding original in-plane slices. Lower is better.}
\label{tab:abdominal_mri_results}
\begin{tabular*}{\tblwidth}{@{} LCC @{}}
\toprule
Method & FID $\downarrow$ & KID $\downarrow$ \\
\midrule
CRIS & \textbf{48.71} & \textbf{0.023} \\
SMORE4 & 66.53 & 0.041 \\
ECLARE & 72.14 & 0.046 \\
SIMPLE & 84.24 & 0.047 \\
Interpolation & 87.87 & 0.068 \\
\bottomrule
\end{tabular*}
\end{table}

\begin{figure*}[t]
\centering
\includegraphics[width=\textwidth]{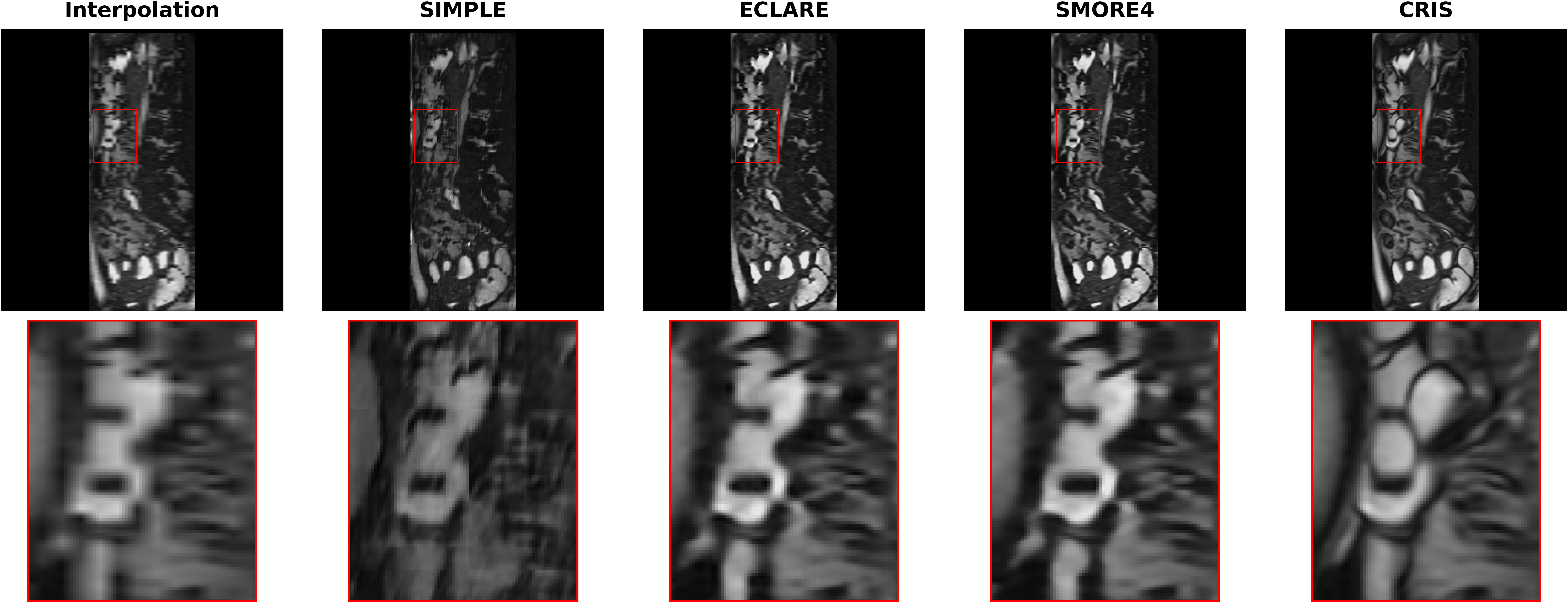}
\vspace{2mm}
\includegraphics[width=\textwidth]{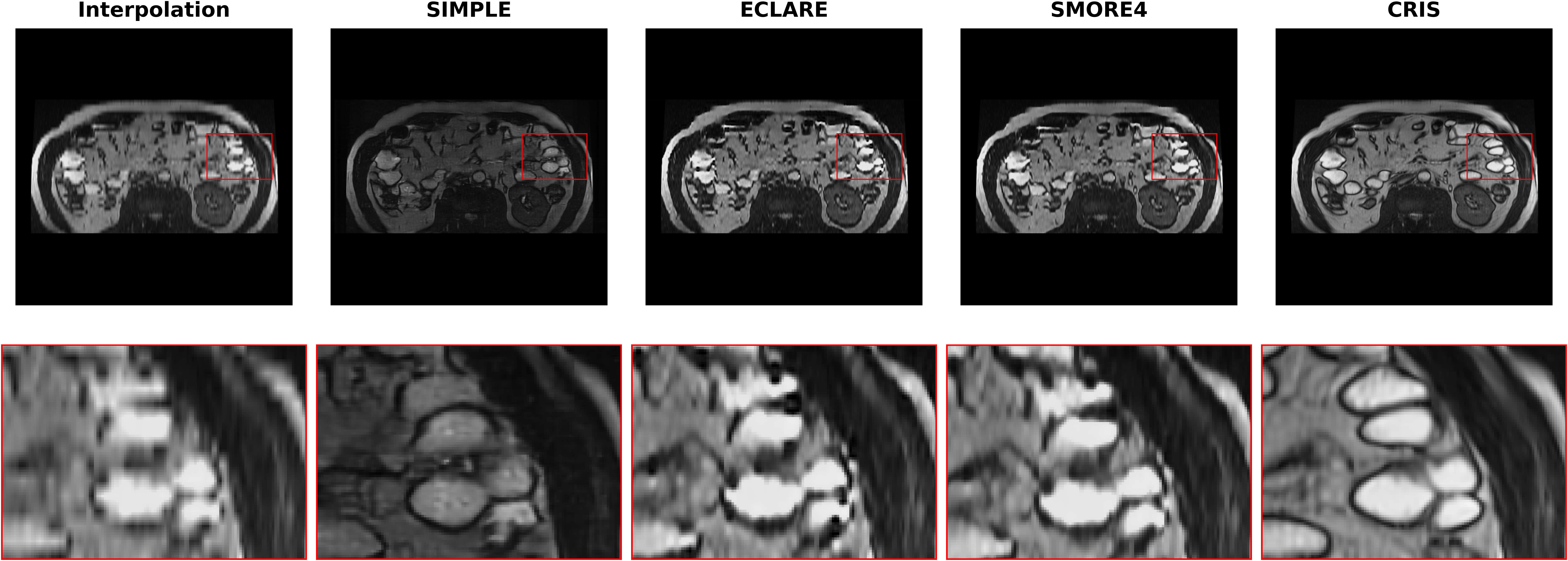}
\caption{Abdominal MRI qualitative comparison showing zoomed-in regions of interest across sagittal and axial planes from a held-out coronal anisotropic source volume with typical gap factor $g\approx6$. Columns are ordered left-to-right as Interpolation, SIMPLE, ECLARE, SMORE4, and CRIS.}
\label{fig:abd_visual_results}
\end{figure*}

Figure~\ref{fig:abd_visual_results} shows qualitative abdominal MRI results. Compared with interpolation, SIMPLE, ECLARE, and SMORE4, CRIS produces more continuous organ boundaries and fewer directional artifacts in sagittal and axial reformats, consistent with the reduction in FID and KID in Table~\ref{tab:abdominal_mri_results}.

\subsection{EPFL FIB-SEM results}
\begin{figure}[pos=t]
\centering
\setlength{\tabcolsep}{2pt}
\renewcommand{\arraystretch}{0.45}
\resizebox{\linewidth}{!}{
\begin{tabular}{c c c c c c c}
& \textbf{Interpolation} & \textbf{NIIV} & \textbf{vEMINR XZ} & \textbf{vEMINR YZ} & \textbf{CRIS} & \textbf{GT} \\

\textbf{$4\times$} &
\includegraphics[angle=90,width=0.10\textwidth]{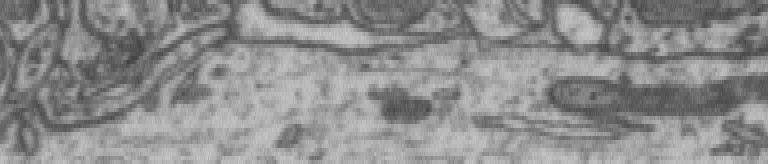} &
\includegraphics[angle=90,width=0.10\textwidth]{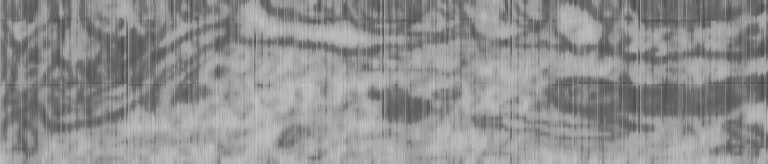} &
\includegraphics[angle=90,width=0.10\textwidth]{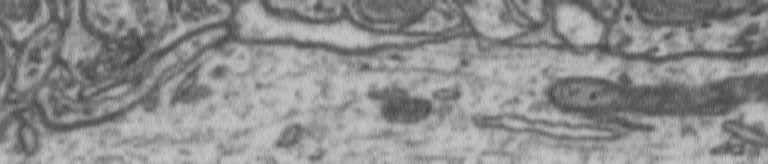} &
\includegraphics[angle=90,width=0.10\textwidth]{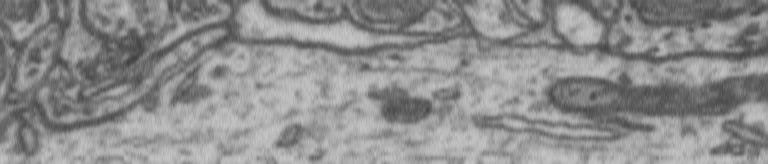} &
\includegraphics[angle=90,width=0.10\textwidth]{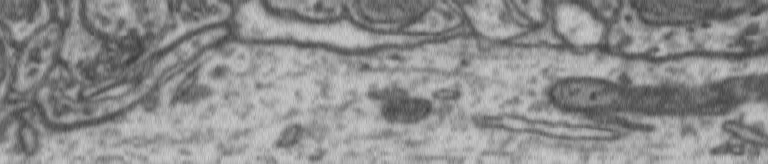} &
\includegraphics[angle=90,width=0.10\textwidth]{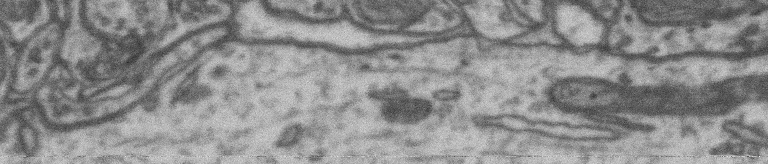} \\

\textbf{$8\times$} &
\includegraphics[angle=90,width=0.10\textwidth]{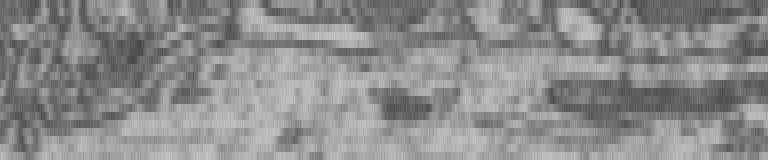} &
\includegraphics[angle=90,width=0.10\textwidth]{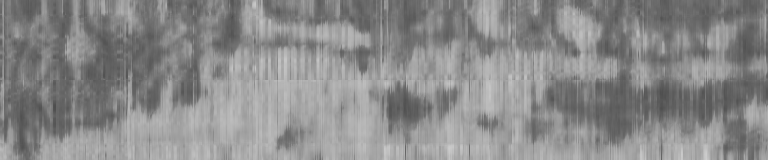} &
\includegraphics[angle=90,width=0.10\textwidth]{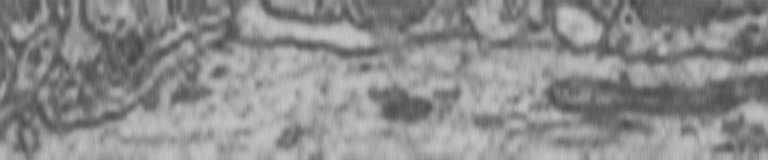} &
\includegraphics[angle=90,width=0.10\textwidth]{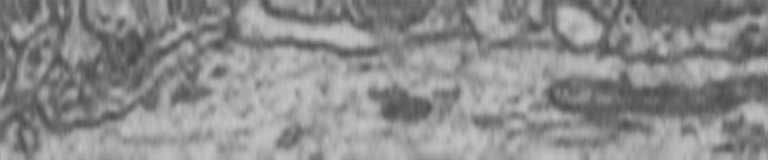} &
\includegraphics[angle=90,width=0.10\textwidth]{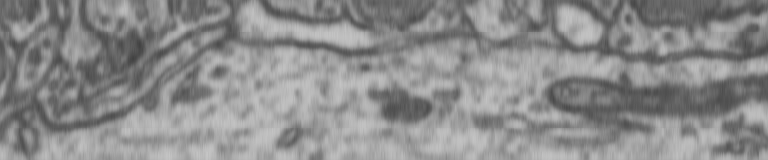} &
\includegraphics[angle=90,width=0.10\textwidth]{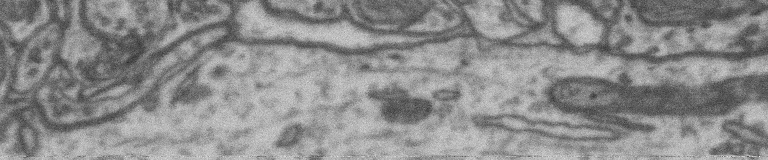} \\
\end{tabular}
}
\caption{Qualitative comparison on the EPFL FIB-SEM dataset in the \textbf{sagittal view} under simulated $4\times$ and $8\times$ through-plane anisotropy. Columns are ordered left-to-right as interpolation, NIIV, vEMINR XZ, vEMINR YZ, CRIS, and isotropic ground truth. CRIS preserves ultrastructural continuity more faithfully under both anisotropy levels, with the advantage becoming more pronounced in the more challenging $8\times$ setting.}
\label{fig:epfl_visual_results}
\end{figure}

Table~\ref{tab:epfl_x4_avg_plane} summarizes average-plane results on the EPFL FIB-SEM benchmark under $4\times$ through-plane anisotropy. CRIS achieves the best performance across all reported metrics, including 3D PSNR, 3D SSIM, GMSD, FID, and KID. It outperforms bicubic interpolation, NIIV, both directional vEMINR reconstructions (XZ and YZ), and the final merged vEMINR volume, indicating that the cross-plane restoration strategy transfers effectively to microscopy when combined with the microscopy-specific degradation and loss design described in Sec.~\ref{sec:modality_specific_details}. The EPFL train and test volumes used in this benchmark each have dimensions $165\times768\times1024$.
Figure~\ref{fig:epfl_visual_results} presents a qualitative comparison on the EPFL benchmark in the sagittal view under both $4\times$ and $8\times$ anisotropy. CRIS yields sharper and more continuous ultrastructural boundaries than interpolation and NIIV, and appears visually more consistent than both directional vEMINR variants, particularly in the more challenging $8\times$ setting.

\begin{table}[pos=t]
\caption{EPFL FIB-SEM average-plane results under simulated $4\times$ through-plane anisotropy ($g=4$). Both train and test subvolumes have dimensions $165\times768\times1024$. Higher is better for 3D PSNR and 3D SSIM; lower is better for GMSD, FID, and KID.}
\label{tab:epfl_x4_avg_plane}
\begin{tabular*}{\tblwidth}{@{} LCCCCC @{}}
\toprule
Method & 3D PSNR $\uparrow$ & 3D SSIM $\uparrow$ & GMSD $\downarrow$ & FID $\downarrow$ & KID $\downarrow$ \\
\midrule
CRIS & \textbf{29.100} & \textbf{0.830} & \textbf{0.0056} & \textbf{37.20} & \textbf{0.033} \\
vEMINR (Final) & 29.099 & 0.826 & 0.0060 & 38.28 & 0.035 \\
vEMINR (XZ) & 29.064 & 0.825 & 0.0060 & 38.17 & 0.034 \\
vEMINR (YZ) & 29.069 & 0.825 & 0.0060 & 37.66 & 0.034 \\
NIIV & 24.824 & 0.664 & 0.0262 & 108.65 & 0.130 \\
Interpolation & 28.853 & 0.812 & 0.0078 & 40.09 & 0.037 \\
\bottomrule
\end{tabular*}
\end{table}

Table~\ref{tab:epfl_x8_avg_plane} reports the more challenging EPFL setting with $8\times$ through-plane anisotropy. Under this severe gap, CRIS achieves the best performance across all reported metrics among the evaluated baselines, with 26.874\,dB 3D PSNR, 0.722 3D SSIM, and the lowest GMSD, FID, and KID. This setting is particularly informative because the larger missing regions amplify hallucination and structural-discontinuity errors, and CRIS remains robust under this stronger anisotropy.

\begin{table}[pos=t]
\caption{EPFL FIB-SEM average-plane results under simulated $8\times$ through-plane anisotropy ($g=8$). Higher is better for 3D PSNR and 3D SSIM; lower is better for GMSD, FID, and KID.}
\label{tab:epfl_x8_avg_plane}
\begin{tabular*}{\tblwidth}{@{} LCCCCC @{}}
\toprule
Method & 3D PSNR $\uparrow$ & 3D SSIM $\uparrow$ & GMSD $\downarrow$ & FID $\downarrow$ & KID $\downarrow$ \\
\midrule
CRIS & \textbf{26.874} & \textbf{0.722} & \textbf{0.0219} & \textbf{58.09} & \textbf{0.058} \\
vEMINR (Final) & 26.660 & 0.704 & 0.0249 & 66.75 & 0.070 \\
vEMINR (XZ) & 26.616 & 0.702 & 0.0249 & 65.57 & 0.068 \\
vEMINR (YZ) & 26.607 & 0.702 & 0.0250 & 64.30 & 0.067 \\
Interpolation & 26.293 & 0.685 & 0.0308 & 79.17 & 0.085 \\
NIIV & 22.819 & 0.489 & 0.0517 & 135.85 & 0.161 \\
\bottomrule
\end{tabular*}
\end{table}

\subsection{FlyEM hemibrain noisy results}
Table~\ref{tab:hemibrain_avg_plane} reports results on the noisy hemibrain benchmark under simulated $8\times$ anisotropy. CRIS improves over NIIV and both interpolation variants across all reported metrics, including CF-PSNR, PSNR, 3D SSIM, GMSD, FID, and KID. For the case-level metrics reported with SD, two-sided paired Wilcoxon signed-rank tests with Holm--Bonferroni correction showed statistically significant CRIS-versus-baseline improvements across all tested comparisons ($p_{\mathrm{adj}}<0.001$). The improvement over NIIV under the same protocol is especially notable because NIIV is a strong self-supervised neural implicit baseline specifically designed for isotropic volume restoration.

\subsection{Microscopy component ablation study}

We further evaluated microscopy-specific CRIS variants on EPFL and FlyEM hemibrain to assess the effect of the loss configuration, rotation augmentation, and adding MRI-style random intensity normalization to the microscopy training pipeline. The ablation results show that no single variant dominates every metric: the final CRIS configuration gives the best or most balanced fidelity and distribution-level performance, while isolated variants can improve individual metrics such as SSIM, GMSD, or KID. Removing microscopy augmentations consistently reduces performance, particularly on hemibrain, whereas adding random intensity normalization is not part of the final microscopy configuration and substantially degrades EPFL $8\times$ fidelity.

\begin{table*}[pos=t]
\caption{Microscopy component ablation study. EPFL results are full-volume average-plane metrics. FlyEM hemibrain PSNR, 3D SSIM, and GMSD are reported as mean $\pm$ SD over 50 test subvolumes; FID and KID are distribution-level metrics and are reported without SD. Higher is better for PSNR and 3D SSIM; lower is better for GMSD, FID, and KID. ``Added random normalization'' denotes a microscopy variant in which MRI-style random intensity normalization was added during training; it is not part of the final microscopy configuration. Bold indicates the best value within each dataset block among CRIS variants.}
\label{tab:microscopy_ablation}
\centering
\scriptsize
\setlength{\tabcolsep}{3pt}
\resizebox{\textwidth}{!}{%
\begin{tabular}{llccccc}
\toprule
Dataset & Variant & PSNR $\uparrow$ & 3D SSIM $\uparrow$ & GMSD $\downarrow$ & FID $\downarrow$ & KID $\downarrow$ \\
\midrule
EPFL $4\times$
& CRIS & \textbf{29.100} & \textbf{0.830} & \textbf{0.0056} & \textbf{37.20} & \textbf{0.033} \\
& All-loss objective & 28.918 & 0.827 & 0.0063 & 41.07 & 0.038 \\
& No augmentations & 28.944 & 0.825 & 0.0065 & 38.74 & 0.036 \\
\midrule
EPFL $8\times$
& CRIS & \textbf{26.874} & 0.722 & 0.0219 & \textbf{58.09} & 0.058 \\
& All-loss objective & 26.800 & \textbf{0.731} & \textbf{0.0202} & 60.54 & 0.061 \\
& No augmentations & 26.760 & 0.720 & 0.0224 & 59.11 & 0.058 \\
& Added random normalization & 23.405 & 0.657 & 0.0380 & 60.02 & \textbf{0.056} \\
\midrule
FlyEM hemibrain $8\times$
& CRIS & \textbf{21.935 $\pm$ 0.437} & 0.696 $\pm$ 0.024 & 0.0447 $\pm$ 0.0048 & \textbf{46.12} & \textbf{0.055} \\
& All-loss objective & 21.922 $\pm$ 0.438 & 0.699 $\pm$ 0.024 & \textbf{0.0439 $\pm$ 0.0047} & 46.33 & 0.056 \\
& No augmentations & 21.794 $\pm$ 0.434 & 0.675 $\pm$ 0.026 & 0.0472 $\pm$ 0.0050 & 50.23 & 0.061 \\
& Added random normalization & 21.876 $\pm$ 0.435 & \textbf{0.702 $\pm$ 0.023} & 0.0440 $\pm$ 0.0046 & 46.53 & 0.056 \\
\bottomrule
\end{tabular}
}
\end{table*}

\subsection{Summary across modalities}
Across both MRI and microscopy, CRIS consistently improves isotropic restoration quality under self-supervised training without paired isotropic targets. In MRI, the strongest gains appear in volumetric fidelity, perceptual quality, and downstream segmentation. In microscopy, CRIS remains robust under both moderate and severe anisotropy and outperforms strong neural implicit baselines under matched degradation protocols. Taken together, these results support the CRIS formulation as a practical route to modality-flexible isotropic restoration through explicit missing-slice modeling, orthogonal stripe completion, and modality-aware self-supervised degradation.

\begin{table*}[pos=t]
\caption{FlyEM hemibrain noisy subvolume results under simulated $8\times$ through-plane anisotropy ($g=8$). The benchmark uses 400 subvolumes total, split in our experiments as 300 training, 50 validation, and 50 test subvolumes; cases 301--350 from the \texttt{hemibrain-volume-noisy-large} training portion were allocated to validation. PSNR, CF-PSNR, 3D SSIM, and GMSD are reported as mean $\pm$ SD over the 50 test subvolumes using the average-plane evaluation. FID and KID are distribution-level metrics and are reported without SD. Higher is better for PSNR, CF-PSNR, and 3D SSIM; lower is better for GMSD, FID, and KID. NN denotes nearest-neighbor interpolation, and Bilinear denotes bilinear interpolation.}
\label{tab:hemibrain_avg_plane}
\centering
\scriptsize
\setlength{\tabcolsep}{3pt}
\resizebox{\textwidth}{!}{%
\begin{tabular}{lcccccc}
\toprule
Method & PSNR $\uparrow$ & CF-PSNR $\uparrow$ & 3D SSIM $\uparrow$ & GMSD $\downarrow$ & FID $\downarrow$ & KID $\downarrow$ \\
\midrule
CRIS
& \textbf{21.935 $\pm$ 0.437}
& \textbf{25.162 $\pm$ 0.680}
& \textbf{0.696 $\pm$ 0.024}
& \textbf{0.0447 $\pm$ 0.0048}
& \textbf{46.12}
& \textbf{0.055} \\

NIIV
& 21.610 $\pm$ 0.459
& 24.575 $\pm$ 0.717
& 0.677 $\pm$ 0.025
& 0.0501 $\pm$ 0.0055
& 48.00
& 0.059 \\

NN
& 20.510 $\pm$ 0.501
& 22.976 $\pm$ 0.722
& 0.626 $\pm$ 0.024
& 0.0758 $\pm$ 0.0078
& 85.11
& 0.109 \\

Bilinear
& 20.960 $\pm$ 0.500
& 23.379 $\pm$ 0.756
& 0.633 $\pm$ 0.027
& 0.0638 $\pm$ 0.0068
& 61.80
& 0.077 \\
\bottomrule
\end{tabular}
}
\end{table*}

\section{Discussion}

CRIS addresses isotropic restoration from anisotropic volumetric imaging without requiring paired isotropic ground truth, a setting that is common in both clinical MRI and volumetric microscopy. By reformulating the problem as cross-plane self-supervised stripe completion, the method leverages the high-resolution in-plane content already available in routine acquisitions and avoids dependence on external supervision. This makes the framework attractive in realistic scenarios where fully isotropic reference data are unavailable or impractical to acquire. In this respect, CRIS extends the practical advantages of self-supervised restoration beyond single-plane formulations previously explored in MRI and microscopy \citep{Zhao2021SMORE,Remedios2023SelfSupSR,Yang2026vEMINR,Troidl2026niiv}.

A central distinction from prior self-supervised MRI super-resolution methods is the explicit modeling of missing data. Whereas methods such as SMORE \citep{Zhao2021SMORE, Remedios2023SelfSupSR} rely on interpolated inputs and learn refinement mappings, CRIS formulates the problem as conditional completion under a known sampling pattern. This difference shifts the learning objective from deblurring to reconstruction, which we find leads to improved volumetric consistency and robustness across acquisition settings.

The results indicate that the proposed cross-plane formulation and the multi-view fusion strategy are complementary. Independent reconstruction of orthogonal reformats captures plane-specific structure effectively, while voxel-wise fusion reduces view-specific artifacts and improves three-dimensional continuity. This is consistent with the qualitative observation that each single-view reconstruction tends to be strongest in its native reconstruction plane, whereas the fused volume provides a better balance for volumetric analysis. More broadly, this supports the idea that explicit cross-plane consistency is a useful inductive bias for isotropic restoration, complementing prior multi-plane and multi-view approaches \citep{Benisty2026SIMPLE,Yang2026vEMINR}.

The dedicated brain MRI robustness study further shows that the learned restoration prior is not tightly tied to a single anisotropy severity or acquisition orientation, as the same trained model remained effective across gap factors $3$--$7$ and across coronal, axial, and sagittal degradations. This is a practical advantage over many learning-based restoration pipelines, where the degradation setting is often baked into the training protocol and fair deployment across different gap factors or acquisition orientations would require additional retraining and tuning. In contrast, CRIS can be trained with a variable-gap self-supervised degradation process and then applied across a range of anisotropy configurations.

A central finding of this study is that a common restoration framework can generalize across modalities when paired with degradation models that reflect the relevant acquisition physics. In MRI, directional Gaussian degradation provides a more suitable approximation of through-plane slice blurring and partial-volume effects, whereas in microscopy, average-pooling-based degradation better matches anisotropic section integration. Thus, the framework is shared, but its degradation operator and loss design remain modality-aware. This distinction appears important for stable self-supervision, since physically mismatched degradations are known to reduce realism and downstream utility in anisotropic restoration pipelines \citep{Remedios2023SelfSupSR,Khateri2025MRISurvey}.

The segmentation results further suggest that the benefits of CRIS extend beyond image appearance. The consistent gains across anatomical structures indicate improved preservation of biologically meaningful morphology, which is critical for downstream medical image analysis. This is especially important because improvements in perceptual or pixel-wise metrics do not necessarily translate into better analysis performance.

Several limitations should be noted. The public brain MRI benchmark is based on synthetic anisotropy, which does not capture all real acquisition artifacts. The abdominal MRI setting is more realistic, but lacks isotropic ground truth for direct full-reference evaluation. In addition, the current fusion stage is deliberately simple and may be suboptimal in regions where the confidence of the orthogonal reconstructions differs substantially. The microscopy experiments, while encouraging, also remain restricted to two benchmark settings and do not yet cover the full diversity of acquisition protocols encountered in large-scale biological imaging.

Future work should investigate adaptive or uncertainty-weighted fusion to reduce the influence of unreliable view-specific predictions, as well as tighter integration with downstream tasks. Broader evaluation across scanners, protocols, anatomies, anisotropy factors, and additional microscopy settings will also be important for establishing the robustness and translational utility of the framework.

Overall, CRIS shows that anisotropic-to-isotropic restoration can be approached as a cross-plane completion problem that reuses a common architecture while adapting the degradation model and training schedule to the imaging modality. Across two MRI cohorts and two microscopy benchmarks, the method consistently improves restoration quality over interpolation and strong self-supervised baselines, while remaining trainable without paired isotropic targets. These findings support cross-plane self-supervision as a practical and modality-flexible strategy for recovering isotropic structure from routine anisotropic volumetric acquisitions.

\FloatBarrier
\section*{Declaration of generative AI and AI-assisted technologies in the manuscript preparation process}

During the preparation of this manuscript, the authors used ChatGPT to assist with drafting parts of the manuscript and writing several code functions used in the research workflow. The tool was not used to generate, alter, or interpret experimental data, results, or figures. All AI-assisted outputs were carefully reviewed, edited, validated, and approved by the authors, who take full responsibility for the final content of the manuscript.

\section*{Data availability}

The public brain MRI data used in this study were drawn from ABIDE and ABIDE-II \citep{DiMartino2014ABIDE,DiMartino2017ABIDEII} and are available through the LONI Image and Data Archive (IDA): \url{https://ida.loni.usc.edu/login.jsp}, subject to the repository's access and data-use requirements. The exact subject identifiers and train/validation/test split used in this work are provided in Supplementary Table~S4.

The EPFL FIB-SEM data are publicly available from the EPFL CVLab Electron Microscopy Dataset page: \url{https://www.epfl.ch/labs/cvlab/data/data-em/}. We used the standard train and test subvolumes described in the manuscript \citep{EPFLDataEM,Lucchi2013WorkingSets}.

The FlyEM hemibrain noisy subvolume data used for the microscopy benchmark can be accessed through the data download link provided in the NIIV GitHub repository: \url{https://github.com/jakobtroidl/niiv-miccai}. In our experiments, cases 301--350 from the \texttt{hemibrain-volume-noisy-large} training portion were allocated to validation, yielding 300 training, 50 validation, and 50 test subvolumes \citep{Troidl2026niiv,NIIVGitHub}.

The private abdominal MRI cohort cannot be made publicly available because it contains clinical imaging data collected under institutional approval and subject to patient privacy and institutional data-use restrictions. Access to these data may be considered by the corresponding author upon reasonable request and subject to institutional approval and data-sharing agreements.

\section*{Code availability}

The code used to train and evaluate CRIS is publicly available at:
\url{https://github.com/adi-hatav/CRIS}.

\section*{Funding}
This work was supported by the Israel Innovation Authority [grant number 90044/90045].

\section*{Declaration of competing interest}

The authors declare that they have no known competing financial interests or personal relationships that could have appeared to influence the work reported in this paper.

\printcredits

\FloatBarrier
\bibliographystyle{cas-model2-names}
\bibliography{references}

\clearpage
\onecolumn

\section*{Supplementary Material}

\setcounter{section}{0}
\setcounter{table}{0}
\setcounter{figure}{0}
\renewcommand{\thesection}{S\arabic{section}}
\renewcommand{\thetable}{S\arabic{table}}
\renewcommand{\thefigure}{S\arabic{figure}}

\setlength{\parskip}{0.25em}
\setlength{\parindent}{0pt}
\setlength{\intextsep}{4pt}
\setlength{\textfloatsep}{4pt}

\Needspace{0.80\textheight}
\section{Per-structure downstream segmentation results}

\Needspace{0.30\textheight}
\refstepcounter{table}\label{tab:supp_s1_segmentation_per_structure}
\noindent\textbf{Table~\thetable.}\par
{\footnotesize Per-structure downstream segmentation consistency across 32 anatomical brain labels on the public brain MRI cohort. Values are reported as mean $\pm$ SD over the eight test volumes for each anatomical label, method, and metric. Higher Dice indicates better overlap, whereas lower ASSD and HD99 indicate better surface agreement. Bold indicates the best value within each anatomical structure and metric group.\par}
\vspace{0.25em}
{\centering
    \tiny
    \setlength{\tabcolsep}{1.35pt}
    \renewcommand{\arraystretch}{0.86}
    \resizebox{\textwidth}{!}{%
    \begin{tabular}{@{}lccccccccc@{}}
        \toprule
        \multirow{2}{*}{Label name} & \multicolumn{3}{c}{Dice $\uparrow$} & \multicolumn{3}{c}{ASSD (mm) $\downarrow$} & \multicolumn{3}{c}{HD99 (mm) $\downarrow$} \\
        \cmidrule(lr){2-4}\cmidrule(lr){5-7}\cmidrule(lr){8-10}
        & CRIS & SMORE4 & Interpolation & CRIS & SMORE4 & Interpolation & CRIS & SMORE4 & Interpolation \\
        \midrule
        left cerebral white matter & \textbf{0.953 $\pm$ 0.009} & 0.940 $\pm$ 0.010 & 0.926 $\pm$ 0.008 & \textbf{0.267 $\pm$ 0.038} & 0.332 $\pm$ 0.040 & 0.412 $\pm$ 0.033 & \textbf{1.421 $\pm$ 0.110} & 1.637 $\pm$ 0.269 & 2.069 $\pm$ 0.123 \\
        left cerebral cortex & \textbf{0.930 $\pm$ 0.008} & 0.913 $\pm$ 0.009 & 0.894 $\pm$ 0.007 & \textbf{0.336 $\pm$ 0.035} & 0.404 $\pm$ 0.034 & 0.475 $\pm$ 0.029 & \textbf{1.421 $\pm$ 0.110} & 1.725 $\pm$ 0.093 & 2.012 $\pm$ 0.107 \\
        left lateral ventricle & \textbf{0.973 $\pm$ 0.006} & 0.964 $\pm$ 0.008 & 0.948 $\pm$ 0.013 & \textbf{0.106 $\pm$ 0.017} & 0.146 $\pm$ 0.030 & 0.228 $\pm$ 0.060 & \textbf{0.977 $\pm$ 0.000} & 1.028 $\pm$ 0.143 & 2.664 $\pm$ 2.387 \\
        left inferior lateral ventricle & \textbf{0.837 $\pm$ 0.050} & 0.794 $\pm$ 0.057 & 0.771 $\pm$ 0.036 & \textbf{0.256 $\pm$ 0.079} & 0.369 $\pm$ 0.121 & 0.407 $\pm$ 0.083 & \textbf{1.677 $\pm$ 0.557} & 2.861 $\pm$ 1.415 & 2.772 $\pm$ 0.629 \\
        left cerebellum white matter & \textbf{0.929 $\pm$ 0.008} & 0.904 $\pm$ 0.014 & 0.894 $\pm$ 0.014 & \textbf{0.280 $\pm$ 0.029} & 0.384 $\pm$ 0.041 & 0.451 $\pm$ 0.051 & \textbf{1.907 $\pm$ 0.263} & 2.740 $\pm$ 0.383 & 3.383 $\pm$ 0.534 \\
        left cerebellum cortex & \textbf{0.958 $\pm$ 0.005} & 0.945 $\pm$ 0.005 & 0.936 $\pm$ 0.005 & \textbf{0.341 $\pm$ 0.033} & 0.442 $\pm$ 0.029 & 0.519 $\pm$ 0.032 & \textbf{1.943 $\pm$ 0.430} & 2.309 $\pm$ 0.206 & 2.692 $\pm$ 0.193 \\
        left thalamus & \textbf{0.968 $\pm$ 0.005} & 0.965 $\pm$ 0.002 & 0.951 $\pm$ 0.004 & \textbf{0.235 $\pm$ 0.033} & 0.263 $\pm$ 0.021 & 0.355 $\pm$ 0.027 & 1.129 $\pm$ 0.209 & \textbf{1.054 $\pm$ 0.152} & 1.531 $\pm$ 0.221 \\
        left caudate & \textbf{0.972 $\pm$ 0.005} & 0.963 $\pm$ 0.004 & 0.948 $\pm$ 0.007 & \textbf{0.133 $\pm$ 0.024} & 0.172 $\pm$ 0.016 & 0.243 $\pm$ 0.032 & \textbf{0.977 $\pm$ 0.000} & \textbf{0.977 $\pm$ 0.000} & 1.078 $\pm$ 0.187 \\
        left putamen & \textbf{0.972 $\pm$ 0.002} & 0.966 $\pm$ 0.003 & 0.956 $\pm$ 0.004 & \textbf{0.169 $\pm$ 0.016} & 0.208 $\pm$ 0.018 & 0.266 $\pm$ 0.026 & \textbf{0.977 $\pm$ 0.000} & 1.028 $\pm$ 0.143 & 1.078 $\pm$ 0.187 \\
        left pallidum & \textbf{0.933 $\pm$ 0.018} & 0.923 $\pm$ 0.017 & 0.907 $\pm$ 0.023 & \textbf{0.313 $\pm$ 0.083} & 0.355 $\pm$ 0.065 & 0.421 $\pm$ 0.099 & \textbf{1.129 $\pm$ 0.209} & 1.301 $\pm$ 0.331 & 1.498 $\pm$ 0.161 \\
        3rd ventricle & \textbf{0.943 $\pm$ 0.012} & 0.920 $\pm$ 0.014 & 0.915 $\pm$ 0.012 & \textbf{0.123 $\pm$ 0.025} & 0.167 $\pm$ 0.026 & 0.174 $\pm$ 0.028 & \textbf{0.977 $\pm$ 0.000} & \textbf{0.977 $\pm$ 0.000} & 1.028 $\pm$ 0.143 \\
        4th ventricle & \textbf{0.928 $\pm$ 0.016} & 0.896 $\pm$ 0.025 & 0.872 $\pm$ 0.034 & \textbf{0.229 $\pm$ 0.044} & 0.365 $\pm$ 0.087 & 0.438 $\pm$ 0.114 & \textbf{1.258 $\pm$ 0.184} & 2.352 $\pm$ 1.941 & 3.236 $\pm$ 2.320 \\
        brain-stem & \textbf{0.975 $\pm$ 0.004} & 0.966 $\pm$ 0.004 & 0.961 $\pm$ 0.004 & \textbf{0.225 $\pm$ 0.031} & 0.300 $\pm$ 0.035 & 0.347 $\pm$ 0.034 & \textbf{1.028 $\pm$ 0.143} & 1.168 $\pm$ 0.280 & 1.453 $\pm$ 0.202 \\
        left hippocampus & \textbf{0.950 $\pm$ 0.007} & 0.936 $\pm$ 0.009 & 0.933 $\pm$ 0.004 & \textbf{0.229 $\pm$ 0.027} & 0.287 $\pm$ 0.045 & 0.304 $\pm$ 0.026 & \textbf{1.028 $\pm$ 0.143} & 1.319 $\pm$ 0.237 & 1.179 $\pm$ 0.216 \\
        left amygdala & \textbf{0.928 $\pm$ 0.009} & 0.905 $\pm$ 0.022 & 0.908 $\pm$ 0.018 & \textbf{0.338 $\pm$ 0.035} & 0.436 $\pm$ 0.097 & 0.430 $\pm$ 0.077 & \textbf{1.179 $\pm$ 0.216} & 1.319 $\pm$ 0.237 & 1.358 $\pm$ 0.271 \\
        CSF & \textbf{0.846 $\pm$ 0.010} & 0.808 $\pm$ 0.014 & 0.767 $\pm$ 0.016 & \textbf{0.358 $\pm$ 0.025} & 0.446 $\pm$ 0.028 & 0.530 $\pm$ 0.035 & \textbf{2.012 $\pm$ 0.107} & 2.355 $\pm$ 0.262 & 2.660 $\pm$ 0.307 \\
        left accumbens area & \textbf{0.932 $\pm$ 0.009} & 0.923 $\pm$ 0.014 & 0.911 $\pm$ 0.018 & \textbf{0.220 $\pm$ 0.028} & 0.250 $\pm$ 0.049 & 0.284 $\pm$ 0.056 & \textbf{1.078 $\pm$ 0.187} & 1.117 $\pm$ 0.272 & \textbf{1.078 $\pm$ 0.187} \\
        left ventral DC & \textbf{0.946 $\pm$ 0.006} & 0.936 $\pm$ 0.005 & 0.930 $\pm$ 0.004 & \textbf{0.256 $\pm$ 0.025} & 0.303 $\pm$ 0.021 & 0.331 $\pm$ 0.020 & \textbf{0.977 $\pm$ 0.000} & 1.179 $\pm$ 0.216 & 1.230 $\pm$ 0.209 \\
        right cerebral white matter & \textbf{0.952 $\pm$ 0.008} & 0.940 $\pm$ 0.008 & 0.926 $\pm$ 0.009 & \textbf{0.266 $\pm$ 0.034} & 0.330 $\pm$ 0.035 & 0.407 $\pm$ 0.036 & \textbf{1.370 $\pm$ 0.192} & 1.599 $\pm$ 0.282 & 2.153 $\pm$ 0.142 \\
        right cerebral cortex & \textbf{0.930 $\pm$ 0.007} & 0.912 $\pm$ 0.007 & 0.894 $\pm$ 0.008 & \textbf{0.336 $\pm$ 0.033} & 0.406 $\pm$ 0.032 & 0.479 $\pm$ 0.032 & \textbf{1.459 $\pm$ 0.144} & 1.647 $\pm$ 0.187 & 1.917 $\pm$ 0.160 \\
        right lateral ventricle & \textbf{0.974 $\pm$ 0.010} & 0.966 $\pm$ 0.011 & 0.951 $\pm$ 0.015 & \textbf{0.105 $\pm$ 0.026} & 0.135 $\pm$ 0.022 & 0.197 $\pm$ 0.029 & \textbf{1.028 $\pm$ 0.143} & 1.104 $\pm$ 0.348 & 1.206 $\pm$ 0.331 \\
        right inferior lateral ventricle & \textbf{0.845 $\pm$ 0.030} & 0.776 $\pm$ 0.056 & 0.762 $\pm$ 0.061 & \textbf{0.260 $\pm$ 0.037} & 0.385 $\pm$ 0.075 & 0.421 $\pm$ 0.074 & \textbf{1.463 $\pm$ 0.399} & 2.130 $\pm$ 0.989 & 2.458 $\pm$ 0.680 \\
        right cerebellum white matter & \textbf{0.932 $\pm$ 0.007} & 0.907 $\pm$ 0.011 & 0.893 $\pm$ 0.016 & \textbf{0.272 $\pm$ 0.029} & 0.398 $\pm$ 0.053 & 0.474 $\pm$ 0.080 & \textbf{1.997 $\pm$ 0.594} & 3.357 $\pm$ 0.693 & 3.886 $\pm$ 0.821 \\
        right cerebellum cortex & \textbf{0.960 $\pm$ 0.004} & 0.947 $\pm$ 0.005 & 0.939 $\pm$ 0.006 & \textbf{0.328 $\pm$ 0.036} & 0.433 $\pm$ 0.037 & 0.498 $\pm$ 0.044 & \textbf{1.770 $\pm$ 0.316} & 2.291 $\pm$ 0.371 & 2.703 $\pm$ 0.328 \\
        right thalamus & \textbf{0.970 $\pm$ 0.005} & 0.963 $\pm$ 0.005 & 0.949 $\pm$ 0.006 & \textbf{0.229 $\pm$ 0.031} & 0.282 $\pm$ 0.036 & 0.381 $\pm$ 0.038 & \textbf{1.028 $\pm$ 0.143} & 1.129 $\pm$ 0.209 & 1.459 $\pm$ 0.144 \\
        right caudate & \textbf{0.973 $\pm$ 0.004} & 0.964 $\pm$ 0.005 & 0.950 $\pm$ 0.005 & \textbf{0.132 $\pm$ 0.023} & 0.171 $\pm$ 0.022 & 0.243 $\pm$ 0.033 & \textbf{0.977 $\pm$ 0.000} & \textbf{0.977 $\pm$ 0.000} & 1.118 $\pm$ 0.197 \\
        right putamen & \textbf{0.974 $\pm$ 0.004} & 0.968 $\pm$ 0.003 & 0.961 $\pm$ 0.004 & \textbf{0.163 $\pm$ 0.030} & 0.202 $\pm$ 0.022 & 0.248 $\pm$ 0.028 & \textbf{0.977 $\pm$ 0.000} & \textbf{0.977 $\pm$ 0.000} & 1.028 $\pm$ 0.143 \\
        right pallidum & \textbf{0.939 $\pm$ 0.022} & 0.932 $\pm$ 0.015 & 0.926 $\pm$ 0.015 & \textbf{0.286 $\pm$ 0.093} & 0.325 $\pm$ 0.068 & 0.348 $\pm$ 0.064 & \textbf{1.078 $\pm$ 0.187} & \textbf{1.078 $\pm$ 0.187} & 1.331 $\pm$ 0.308 \\
        right hippocampus & \textbf{0.942 $\pm$ 0.017} & 0.926 $\pm$ 0.020 & 0.922 $\pm$ 0.023 & \textbf{0.260 $\pm$ 0.076} & 0.325 $\pm$ 0.081 & 0.343 $\pm$ 0.095 & \textbf{1.078 $\pm$ 0.187} & 1.257 $\pm$ 0.321 & 1.356 $\pm$ 0.464 \\
        right amygdala & \textbf{0.923 $\pm$ 0.018} & 0.915 $\pm$ 0.015 & 0.915 $\pm$ 0.013 & \textbf{0.371 $\pm$ 0.081} & 0.408 $\pm$ 0.071 & 0.407 $\pm$ 0.052 & 1.542 $\pm$ 0.440 & \textbf{1.409 $\pm$ 0.224} & 1.430 $\pm$ 0.344 \\
        right accumbens area & \textbf{0.943 $\pm$ 0.009} & 0.926 $\pm$ 0.012 & 0.916 $\pm$ 0.014 & \textbf{0.186 $\pm$ 0.028} & 0.239 $\pm$ 0.033 & 0.279 $\pm$ 0.046 & \textbf{0.977 $\pm$ 0.000} & 1.028 $\pm$ 0.143 & 1.129 $\pm$ 0.209 \\
        right ventral DC & \textbf{0.946 $\pm$ 0.005} & 0.936 $\pm$ 0.007 & 0.933 $\pm$ 0.009 & \textbf{0.249 $\pm$ 0.021} & 0.293 $\pm$ 0.034 & 0.309 $\pm$ 0.039 & \textbf{0.977 $\pm$ 0.000} & 1.179 $\pm$ 0.216 & 1.230 $\pm$ 0.209 \\
        \bottomrule
    \end{tabular}%
    }
\par}
\vspace{0.9em}

\Needspace{0.35\textheight}
\section{Dataset-specific CRIS configurations and selected hyperparameters}

Table~\ref{tab:supp_s2_configs_hparams} reports the dataset-specific CRIS configurations and selected hyperparameter values for the four datasets used in the manuscript.

\Needspace{0.30\textheight}
\refstepcounter{table}\label{tab:supp_s2_configs_hparams}
\noindent\textbf{Table~\thetable.}\par
{\footnotesize Dataset-specific CRIS configurations and selected hyperparameters.\par}
\vspace{0.25em}
{\centering
\scriptsize
\renewcommand{\arraystretch}{1.02}
\begin{tabularx}{\linewidth}{@{}p{0.23\linewidth}p{0.18\linewidth}p{0.18\linewidth}p{0.18\linewidth}Y@{}}
\toprule
Hyperparameter & Brain MRI & Abdominal MRI & EPFL FIB-SEM & FlyEM hemibrain noisy \\
\midrule
Gap used during training & $5$ & $6$ with optional variation to $7$ & Separate runs for gaps $4$ and $8$ & $8$ \\
Patch dimensions (Y $\times$ X) & $256 \times 256$ & $512 \times 512$ & $128 \times 128$ & $128 \times 128$ \\
Learning rate & $1\times10^{-4}$ & $5\times10^{-5}$ & $5\times10^{-5}$ & $5\times10^{-5}$ \\
Batch size & $24$ & $16$ & $24$ & $64$ \\
Base filters & $118$ & $96$ & $118$ & $118$ \\
Patience & $15$ & $15$ & $15$ & $15$ \\
Degradation operator & Directional 1D Gaussian blur + periodic masking & Directional 1D Gaussian blur + periodic masking & Average pooling + periodic masking & Average pooling + periodic masking \\
Gaussian $\sigma$ / average-pooling kernel & $0.75\sqrt{\mathrm{gap}}$ & $0$ & gap & gap \\
SSIM weight & $5$ & $5$ & $0.5$ & $0.5$ \\
Sobel weight & $5$ & $5$ & Not used & Not used \\
Focal-frequency weight & $10$ & $10$ & Not used & Not used \\
\bottomrule
\end{tabularx}
\par}
\vspace{0.9em}

\Needspace{0.45\textheight}
\section{Baseline implementation details}
\label{sec:supp_baseline_implementation}

All baseline methods were evaluated using the same train, validation, and test partitions described in the manuscript. Held-out isotropic test references were used only for final evaluation in datasets where such references exist, and were not used for training, validation, hyperparameter selection, or model selection. When official implementations were available, we used the supplied code and changed only the data-format interface, path configuration, spacing or gap-factor settings, and output conversion required to evaluate the method on our datasets.

\paragraph{SMORE4.}
SMORE4 was run using the supplied implementation. In the brain MRI benchmark, it was optimized separately for each held-out coronal test volume, following the internal-learning setting of the method. The input to SMORE4 was the same anisotropic coronal source volume used for CRIS inference. This setting ensures that the comparison uses the same held-out source orientation and that no held-out isotropic test reference is used during optimization.

\paragraph{ECLARE.}
ECLARE was evaluated using the authors' supplied implementation (\url{https://github.com/sremedios/eclare}). It was run on the identical held-out anisotropic test volumes matching our exact data partitions and evaluation inputs to ensure a fair comparison.

\paragraph{SIMPLE.}
For abdominal MRI, SIMPLE was evaluated using the supplied code and data configuration. For the public brain MRI cohort, we used the supplied SIMPLE implementation with minor configuration adaptations required for the brain MRI data format and spacing. Brain MRI is not the original target domain of SIMPLE, and the corresponding results should therefore be interpreted as a cross-domain application of the supplied method rather than as an optimized domain-specific SIMPLE model.

\paragraph{SA-INR.}
SA-INR is originally formulated for a supervised or paired reconstruction setting. Because paired isotropic ground-truth volumes are not available during training in our self-supervised task, we used the supplied SA-INR code in an adapted setting. The training and validation data supplied to SA-INR were anisotropic coronal brain MRI volumes rather than paired isotropic target volumes. A separate SA-INR network was trained for each held-out test case. No held-out isotropic test reference was used during SA-INR optimization.

\paragraph{ATME.}
ATME was trained separately for each degradation plane using paired 2D samples derived from the anisotropic training volumes. The input was not an isotropic ground-truth volume. Instead, it was generated from the anisotropic volume by linear resampling or interpolation to the target grid and by applying the plane-dependent stride pattern. The target for each training pair was the corresponding real acquired slice from the original anisotropic volume. The model was optimized as a conditional image-to-image translation baseline using an adversarial loss with a PatchGAN discriminator and an L1 reconstruction loss.

\paragraph{NIIV.}
NIIV was run using the authors' supplied implementation. For the EPFL FIB-SEM benchmark, the degraded training and testing TIFF volumes were converted into the NumPy directory structure expected by the NIIV code. The model code was not modified. During reconstruction, NIIV produced patch-wise outputs. These outputs were loaded from the evaluation directories, the padded through-plane slices were removed according to the detected output-to-input depth ratio, and the 48 spatial patches were tiled back into the full $Z\times768\times1024$ EPFL volume before evaluation. This reconstruction step was used only to convert NIIV's patch outputs into the full-volume format required by our metric pipeline.

\paragraph{vEMINR.}
vEMINR was evaluated using the authors' supplied implementation on the degraded EPFL volumes. We report the directional XZ and YZ reconstructions as well as the merged output when provided by the method. This preserves the intended directional reconstruction behavior of vEMINR while allowing comparison against the fused CRIS output on the same held-out EPFL test volume.

\Needspace{0.78\textheight}
\section{Brain MRI robustness across gap factors and degradation planes}

To complement the main brain MRI benchmark, we trained one additional CRIS model using the public brain MRI training split with randomized training-time gap factors $g\in\{3,4,5,6,7\}$ in the self-supervised degradation process. This single model was then evaluated, without retraining or configuration-specific tuning, across synthetic gap factors $g\in\{3,4,5,6,7\}$ and across coronal, axial, and sagittal degradation planes. Supplementary Table~S3 reports the quantitative results for each configuration using plane-averaged metrics, and Supplementary Fig.~S1 shows representative qualitative cross-plane comparisons for a test case.

\Needspace{0.30\textheight}
\refstepcounter{table}\label{tab:supp_s3_brain_robustness}
\noindent\textbf{Table~\thetable.}\par
{\footnotesize Brain MRI robustness results across degradation planes and gap factors for the public multi-site cohort. PSNR, SSIM 3D, GMSD, and Edges 3D are reported as mean~$\pm$~SD over the eight test cases, after averaging the three evaluation planes within each case. Higher is better for PSNR, SSIM 3D, and Edges 3D; lower is better for GMSD, FID, and KID. Bold indicates the best mean value within each configuration.\par}
\vspace{0.25em}
{\centering
\scriptsize
\setlength{\tabcolsep}{1.35pt}
\renewcommand{\arraystretch}{0.88}
\begin{tabular}{@{}lllcccccc@{}}
\toprule
Config plane & Gap & Method & PSNR $\uparrow$ & SSIM 3D $\uparrow$ & GMSD $\downarrow$ & Edges 3D $\uparrow$ & FID $\downarrow$ & KID $\downarrow$ \\
\midrule
coronal & 3 & CRIS & \textbf{37.584} $\pm$ 0.429 & \textbf{0.984} $\pm$ 0.001 & \textbf{0.0051} $\pm$ 0.0003 & \textbf{0.982} $\pm$ 0.001 & \textbf{21.28} & \textbf{0.0048} \\
coronal & 3 & Interpolation & 33.710 $\pm$ 0.495 & 0.962 $\pm$ 0.003 & 0.0175 $\pm$ 0.0010 & 0.942 $\pm$ 0.004 & 51.73 & 0.0272 \\
coronal & 4 & CRIS & \textbf{35.238} $\pm$ 0.434 & \textbf{0.974} $\pm$ 0.002 & \textbf{0.0102} $\pm$ 0.0005 & \textbf{0.966} $\pm$ 0.003 & \textbf{28.38} & \textbf{0.0095} \\
coronal & 4 & Interpolation & 31.180 $\pm$ 0.493 & 0.935 $\pm$ 0.005 & 0.0312 $\pm$ 0.0015 & 0.903 $\pm$ 0.006 & 77.83 & 0.0474 \\
coronal & 5 & CRIS & \textbf{33.547} $\pm$ 0.441 & \textbf{0.962} $\pm$ 0.003 & \textbf{0.0154} $\pm$ 0.0007 & \textbf{0.948} $\pm$ 0.004 & \textbf{36.16} & \textbf{0.0141} \\
coronal & 5 & Interpolation & 29.488 $\pm$ 0.469 & 0.908 $\pm$ 0.007 & 0.0429 $\pm$ 0.0019 & 0.867 $\pm$ 0.009 & 102.06 & 0.0671 \\
coronal & 6 & CRIS & \textbf{32.580} $\pm$ 0.455 & \textbf{0.953} $\pm$ 0.004 & \textbf{0.0201} $\pm$ 0.0009 & \textbf{0.933} $\pm$ 0.005 & \textbf{43.24} & \textbf{0.0193} \\
coronal & 6 & Interpolation & 28.753 $\pm$ 0.511 & 0.887 $\pm$ 0.008 & 0.0522 $\pm$ 0.0025 & 0.841 $\pm$ 0.010 & 119.03 & 0.0816 \\
coronal & 7 & CRIS & \textbf{31.266} $\pm$ 0.478 & \textbf{0.940} $\pm$ 0.005 & \textbf{0.0257} $\pm$ 0.0011 & \textbf{0.915} $\pm$ 0.007 & \textbf{47.89} & \textbf{0.0199} \\
coronal & 7 & Interpolation & 27.664 $\pm$ 0.513 & 0.865 $\pm$ 0.009 & 0.0612 $\pm$ 0.0032 & 0.815 $\pm$ 0.011 & 119.65 & 0.0756 \\
\midrule
axial & 3 & CRIS & \textbf{35.968} $\pm$ 0.427 & \textbf{0.976} $\pm$ 0.002 & \textbf{0.0081} $\pm$ 0.0004 & \textbf{0.971} $\pm$ 0.002 & \textbf{31.15} & \textbf{0.0115} \\
axial & 3 & Interpolation & 33.114 $\pm$ 0.485 & 0.949 $\pm$ 0.004 & 0.0213 $\pm$ 0.0016 & 0.926 $\pm$ 0.006 & 61.14 & 0.0345 \\
axial & 4 & CRIS & \textbf{34.323} $\pm$ 0.434 & \textbf{0.964} $\pm$ 0.003 & \textbf{0.0140} $\pm$ 0.0007 & \textbf{0.952} $\pm$ 0.004 & \textbf{35.08} & \textbf{0.0148} \\
axial & 4 & Interpolation & 31.169 $\pm$ 0.480 & 0.921 $\pm$ 0.005 & 0.0347 $\pm$ 0.0025 & 0.886 $\pm$ 0.009 & 73.66 & 0.0449 \\
axial & 5 & CRIS & \textbf{32.933} $\pm$ 0.403 & \textbf{0.952} $\pm$ 0.003 & \textbf{0.0194} $\pm$ 0.0008 & \textbf{0.934} $\pm$ 0.006 & \textbf{46.23} & \textbf{0.0211} \\
axial & 5 & Interpolation & 29.767 $\pm$ 0.445 & 0.894 $\pm$ 0.007 & 0.0448 $\pm$ 0.0027 & 0.854 $\pm$ 0.010 & 106.57 & 0.0700 \\
axial & 6 & CRIS & \textbf{32.022} $\pm$ 0.424 & \textbf{0.941} $\pm$ 0.004 & \textbf{0.0246} $\pm$ 0.0012 & \textbf{0.917} $\pm$ 0.007 & \textbf{44.85} & \textbf{0.0208} \\
axial & 6 & Interpolation & 29.059 $\pm$ 0.454 & 0.876 $\pm$ 0.008 & 0.0532 $\pm$ 0.0030 & 0.830 $\pm$ 0.012 & 106.42 & 0.0699 \\
axial & 7 & CRIS & \textbf{31.203} $\pm$ 0.404 & \textbf{0.930} $\pm$ 0.006 & \textbf{0.0296} $\pm$ 0.0014 & \textbf{0.900} $\pm$ 0.008 & \textbf{50.10} & \textbf{0.0242} \\
axial & 7 & Interpolation & 28.410 $\pm$ 0.441 & 0.858 $\pm$ 0.009 & 0.0601 $\pm$ 0.0031 & 0.812 $\pm$ 0.012 & 117.53 & 0.0792 \\
\midrule
sagittal & 3 & CRIS & \textbf{35.513} $\pm$ 0.405 & \textbf{0.972} $\pm$ 0.002 & \textbf{0.0093} $\pm$ 0.0005 & \textbf{0.965} $\pm$ 0.003 & \textbf{36.01} & \textbf{0.0132} \\
sagittal & 3 & Interpolation & 32.391 $\pm$ 0.469 & 0.942 $\pm$ 0.004 & 0.0228 $\pm$ 0.0013 & 0.911 $\pm$ 0.005 & 77.91 & 0.0434 \\
sagittal & 4 & CRIS & \textbf{33.843} $\pm$ 0.439 & \textbf{0.960} $\pm$ 0.003 & \textbf{0.0150} $\pm$ 0.0009 & \textbf{0.946} $\pm$ 0.005 & \textbf{54.53} & \textbf{0.0256} \\
sagittal & 4 & Interpolation & 30.496 $\pm$ 0.482 & 0.911 $\pm$ 0.006 & 0.0368 $\pm$ 0.0020 & 0.868 $\pm$ 0.008 & 124.78 & 0.0825 \\
sagittal & 5 & CRIS & \textbf{32.629} $\pm$ 0.435 & \textbf{0.948} $\pm$ 0.004 & \textbf{0.0202} $\pm$ 0.0009 & \textbf{0.928} $\pm$ 0.006 & \textbf{52.68} & \textbf{0.0238} \\
sagittal & 5 & Interpolation & 29.120 $\pm$ 0.452 & 0.883 $\pm$ 0.008 & 0.0485 $\pm$ 0.0026 & 0.833 $\pm$ 0.011 & 119.95 & 0.0765 \\
sagittal & 6 & CRIS & \textbf{31.776} $\pm$ 0.464 & \textbf{0.937} $\pm$ 0.005 & \textbf{0.0253} $\pm$ 0.0011 & \textbf{0.911} $\pm$ 0.007 & \textbf{71.84} & \textbf{0.0380} \\
sagittal & 6 & Interpolation & 28.187 $\pm$ 0.492 & 0.859 $\pm$ 0.008 & 0.0584 $\pm$ 0.0034 & 0.806 $\pm$ 0.011 & 168.09 & 0.1211 \\
sagittal & 7 & CRIS & \textbf{30.965} $\pm$ 0.456 & \textbf{0.926} $\pm$ 0.005 & \textbf{0.0297} $\pm$ 0.0011 & \textbf{0.896} $\pm$ 0.008 & \textbf{80.24} & \textbf{0.0441} \\
sagittal & 7 & Interpolation & 27.491 $\pm$ 0.486 & 0.837 $\pm$ 0.009 & 0.0665 $\pm$ 0.0036 & 0.785 $\pm$ 0.013 & 182.48 & 0.1357 \\
\bottomrule
\end{tabular}
\par}
\vspace{0.9em}

\Needspace{0.62\textheight}
{\centering
\includegraphics[width=0.80\linewidth]{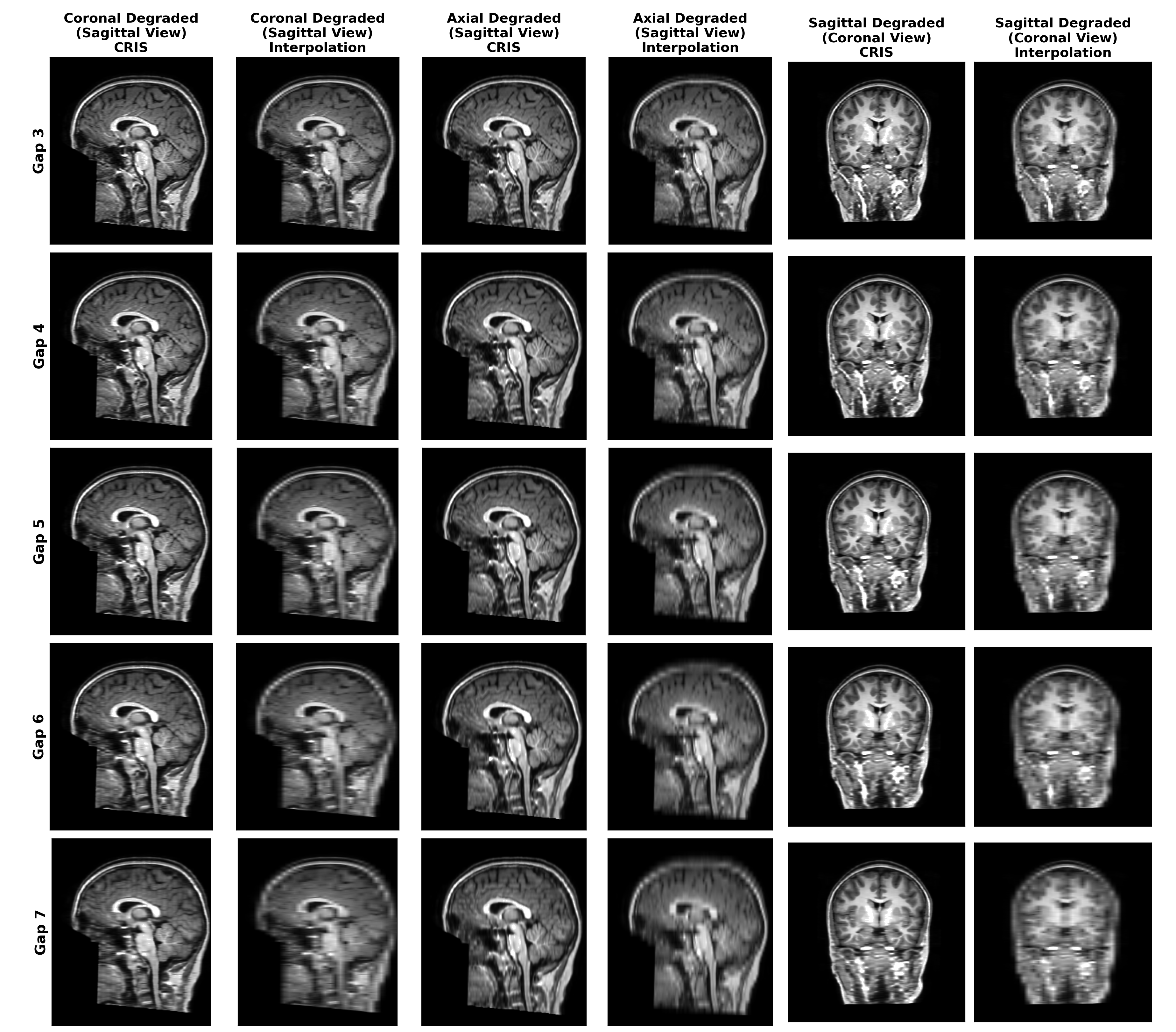}
\par}
\refstepcounter{figure}\label{fig:supp_brain_gap_qualitative}
\noindent\textbf{Figure~\thefigure.}\par
{\footnotesize Representative qualitative comparison of CRIS and interpolation across gap factors and degradation directions on the public brain MRI cohort. Rows correspond to gap factors $g=3$ to $g=7$. Columns show representative orthogonal cross-plane views for coronal-degraded, axial-degraded, and sagittal-degraded inputs, each reconstructed by CRIS and interpolation. For coronal and axial degradations, sagittal views are shown; for sagittal degradation, coronal views are shown. CRIS more effectively suppresses through-plane blur and staircasing artifacts across all tested configurations.\par}
\vspace{0.9em}

\Needspace{0.35\textheight}
\section{Brain MRI subject split}

Supplementary Table~\ref{tab:supp_s4_brain_mri_subject_split} lists the 63 ABIDE/ABIDE-II subject identifiers used for the public brain MRI dataset split. The raw imaging data are available through the LONI Image and Data Archive (IDA), subject to the repository's access and data-use requirements.

\Needspace{0.30\textheight}
\refstepcounter{table}\label{tab:supp_s4_brain_mri_subject_split}
\noindent\textbf{Table~\thetable.}\par
{\footnotesize Subject identifiers used for the public brain MRI dataset split. $n=63$ total: 48 training, 7 validation, and 8 testing.\par}
\vspace{0.25em}
{\centering
\scriptsize
\setlength{\tabcolsep}{3pt}
\renewcommand{\arraystretch}{1.02}
\begin{tabularx}{\linewidth}{@{}l c Y@{}}
\toprule
Dataset split & $n$ & Subject IDs \\
\midrule
Training & 48 & 50682, 50683, 50685, 50689, 50690, 50691, 50692, 50693, 50694, 50695, 50696, 50697, 50698, 50699, 50700, 50701, 50702, 50703, 50704, 50707, 50709, 50710, 50711, 50725, 50726, 50727, 50728, 50730, 50732, 50733, 50734, 50735, 50736, 50738, 50739, 50740, 50741, 50745, 50746, 50747, 50748, 50749, 50750, 50751, 50752, 50753, 50754, 50755 \\
Validation & 7 & 50686, 50688, 50705, 50708, 50737, 50742, 50756 \\
Test & 8 & 50687, 50722, 50723, 50724, 50731, 50743, 50744, 50757 \\
\bottomrule
\end{tabularx}
\par}
\vspace{0.9em}

\end{document}